\documentclass[10pt,journal,compsoc]{IEEEtran}

\usepackage{amsmath,amsfonts,bm}

\def\eqref#1{equation~\ref{#1}}
\def\1{\bm{1}}

\def\vb{{\bm{b}}}

\def\vw{{\bm{w}}}
\def\vx{{\bm{x}}}
\def\vy{{\bm{y}}}

\DeclareMathAlphabet{\mathsfit}{\encodingdefault}{\sfdefault}{m}{sl}
\SetMathAlphabet{\mathsfit}{bold}{\encodingdefault}{\sfdefault}{bx}{n}

\DeclareMathOperator*{\argmax}{arg\,max}

\makeatletter
\let\MYcaption\@makecaption
\makeatother
\usepackage[font=footnotesize]{subcaption}
\makeatletter
\let\@makecaption\MYcaption
\makeatother

\usepackage{url}

\usepackage{graphicx}
\graphicspath{{figure/}}

\usepackage[american]{babel}
\usepackage{microtype}
\usepackage{multirow}
\usepackage{booktabs}

\usepackage[cmyk]{xcolor}

\newcommand{\PractiseNoSpace}{\textsc{Practise}} % This one does NOT produce a space after it
\newcommand{\Practise}{\textsc{Practise} } % This one produces a space after it
\newtheorem{conjecture}{Conjecture}
\newtheorem{claim}{Claim}

\makeatletter
\newif\if@restonecol
\makeatother

\usepackage[ruled,vlined]{algorithm2e}
\usepackage{algpseudocode}

\allowdisplaybreaks[3]

\ifCLASSOPTIONcompsoc
  \usepackage[nocompress]{cite}
\else
  \usepackage{cite}
\fi

\begin{document}

\title{Practical Network Acceleration with Tiny Sets: Hypothesis, Theory, and Algorithm}

\author{Guo-Hua~Wang,
	and~Jianxin~Wu,~\IEEEmembership{Member,~IEEE}% <-this % stops a space
	\IEEEcompsocitemizethanks{\IEEEcompsocthanksitem G.-H. Wang and J. Wu are with the State Key Laboratory for Novel Software Technology, Nanjing University, Nanjing 210023, China. J. Wu is the corresponding author. This paper was partly supported by the National Natural Science Foundation of China under Grant
	62276123 and Grant 61921006.\protect\\
	E-mail: wangguohua@lamda.nju.edu.cn, wujx2001@gmail.com
}}

\markboth{Submitted to IEEE Trans. PAMI}
{Practical Network Acceleration with Tiny Sets: Hypothesis, Theory, and Algorithm}

\IEEEtitleabstractindextext{%
\begin{abstract}
  Due to data privacy issues, accelerating networks with tiny training sets has become a critical need in practice. Previous methods achieved promising results empirically by filter-level pruning. In this paper, we both study this problem theoretically and propose an effective algorithm aligning well with our theoretical results. First, we propose the finetune convexity hypothesis to explain why recent few-shot compression algorithms do not suffer from overfitting problems. Based on it, a theory is further established to explain these methods for the first time. Compared to naively finetuning a pruned network, feature mimicking is proved to achieve a lower variance of parameters and hence enjoys easier optimization. With our theoretical conclusions, we claim dropping blocks is a fundamentally superior few-shot compression scheme in terms of more convex optimization and a higher acceleration ratio. To choose which blocks to drop, we propose a new metric, recoverability, to effectively measure the difficulty of recovering the compressed network. Finally, we propose an algorithm named \Practise to accelerate networks using only tiny training sets. \Practise outperforms previous methods by a significant margin. For 22\% latency reduction, it surpasses previous methods by on average 7 percentage points on ImageNet-1k. It also works well under data-free or out-of-domain data settings. Our code is at \url{https://github.com/DoctorKey/Practise}. 
\end{abstract}

}

\maketitle

\IEEEdisplaynontitleabstractindextext

\IEEEpeerreviewmaketitle

\ifCLASSOPTIONcompsoc
\IEEEraisesectionheading{\section{Introduction}\label{sec:introduction}}
\else
\section{Introduction}
\fi

\IEEEPARstart{D}{eep} networks have achieved remarkable success, but suffer from high computational costs. To accelerate them, many methods have been proposed, such as network pruning~\cite{L1_norm,thinet,slimming,JointPruning}, decoupling~\cite{DW_PW,LRD} and quantization~\cite{quantization,bannerscalable}. However, most previous methods rely on \emph{the original training set} (i.e., all the training data) to recover the compressed model's accuracy. But, to preserve data privacy and/or to achieve fast deployment, only scarce training data may be available. For example, a customer often asks an algorithmic provider to accelerate their networks, but the whole training data \emph{cannot} be made available due to privacy concerns. Only the uncompressed model and a few training examples are provided. In some extreme cases, not even a single data point is accessible. Algorithmic engineers need to synthesize images or collect some out-of-domain training images by themselves. Hence, to accelerate a deep network with only very few samples is emerging as a critical problem.

Previous work~\cite{FSKD,CD,MiR} have achieved promising results. However, most of them address few-shot compression by empirically designing algorithms but leave many fundamental questions unanswered, such as 
\begin{itemize}
	\item Why is it possible to finetune a pruned model with very few samples without suffering from overfitting?
	\item Why must knowledge distillation (KD) especially feature-based mimicking be adopted in finetuning? What benefits does KD or feature mimicking enjoy?
	\item Which compression scheme is best in few-shot compression? What properties should one compression scheme possess to achieve better performance?
\end{itemize}
This paper aims at answering all these keys questions from both theoretical and empirical perspectives.

Training a deep network from scratch is highly non-convex and leads to a complex optimization landscape~\cite{draxler2018essentially,frankle2020linear,git_rebasin}. But we find that finetuning a pruned model is surprisingly convex. Based on lots of experiments, we propose a \emph{finetune convexity hypothesis}, which surmises that the loss landscape of finetuning a pruned model is convex w.r.t. the parameters. Furthermore, the degree of convexity decides the amount of training data required during finetuning. The convexity property allows few-shot finetuning of the pruned model without serious overfitting. To better understand why this convexity occurs, we reformulate the model compression process to insert linear adaptor modules around pruned structures and \emph{only} finetune these modules with few examples. Experiments show that its optimization landscape and accuracy are almost the same as those of end-to-end finetuning. That is, \emph{the difference between the original model and the pruned model (before finetuning) is convex or even close to linear in the latent representation/feature space}.

Although the convexity property reduces few-shot finetuning's difficulty, there are still large accuracy gaps between different few-shot compression methods. Compared to vanilla finetuning, existing work find that knowledge distillation is effective in practice, especially feature mimicking (let the student mimic the penultimate feature of the teacher~\cite{MiR}). However, this crucial empirical finding lacks a theoretical foundation. On top of our finetune convexity hypothesis, we establish a theory for feature mimicking, which studies the variance of learned parameters. We prove that compared to vanilla finetuning, parameters will achieve lower variance via feature mimicking . Hence, feature mimicking makes the final model more stable. To the best of our knowledge, it is the \emph{first theory to successfully show why feature mimicking outperforms vanilla finetuning}.

\begin{figure}
	\centering
	\includegraphics[width=0.65\linewidth]{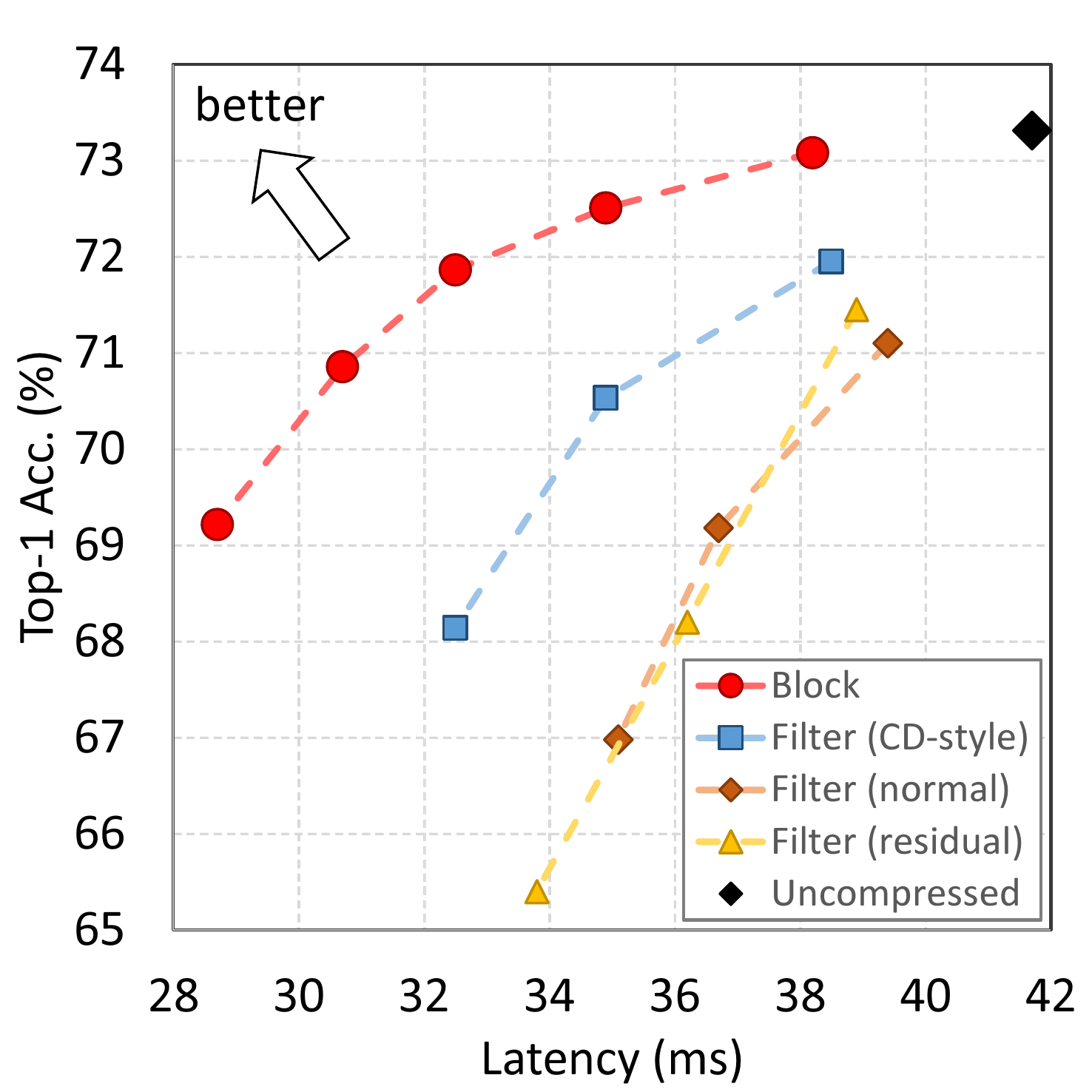}
	\caption{Comparison of different structure compression schemes with 500 training images. Our scheme (`Block') outperforms previous schemes dominantly for the latency-accuracy tradeoff. The ResNet-34 model was compressed on ImageNet-1k and tested on an NVIDIA TITAN Xp GPU.}
	\label{fig:top1-latency}
\end{figure}

The next crucial problem is how to prune the network structure. Almost all previous few-shot compression work~\cite{FSKD,CD,MiR} adopted filter-level pruning. In this paper, we are the first  to reveal that block-level pruning is fundamentally superior in few-shot compression. First, pruning blocks enjoys \emph{a more convex optimization landscape}, which requires fewer data compared to pruning filters. Second, it \emph{achieves much faster speed on real-world computing devices} than filter pruning when the FLOPs are the same. Fig.~\ref{fig:top1-latency} shows that dropping blocks dominantly outperforms previous compression schemes for the latency vs. accuracy tradeoff. Table~\ref{tab:drop-block-BP} further reports that \emph{an embarrassingly simple dropping block baseline (i.e., finetune without any other processing) already surpasses existing methods which use complicated techniques}. The baseline, `BP (blocks)', simply removes the first few blocks and finetunes the pruned network with only the cross-entropy loss.

\begin{table}
  \caption{Top-1 validation accuracy (\%) on ImageNet-1k for different compression schemes. ResNet-34 was accelerated by reducing $16\%$ latency with 50 training images. Previous methods prune filters with the `normal' style. Our block-level scheme, `BP (blocks)', simply removes the first few blocks and finetune the pruned network by back propagation.}
  \label{tab:drop-block-BP}
	\centering
  \small
	\begin{tabular}{ccccc}
		\toprule
    KD~\cite{KD} & FSKD~\cite{FSKD}	& CD~\cite{CD} & MiR~\cite{MiR} & BP (blocks)	\\
		\midrule
    44.5 & 45.3 & 56.2 & 64.1 & \textbf{66.5} \\
		\bottomrule
	\end{tabular}
\end{table}

To achieve better block pruning, we study the strategy to choose which blocks to drop with only scarce training samples. Several criteria~\cite{DBP,e_ResNet,CURL} have been proposed for pruning blocks, but some~\cite{DBP,e_ResNet} require a large amount of data and  others~\cite{CURL} only evaluate the output difference before/after block removal. We notice that \emph{although dropping some blocks significantly changes the feature maps, they are easily recovered by a finetuning even with a tiny training set,} because of the finetune convexity hypothesis. Simply measuring the difference between pruned/original networks is at best suboptimal. We need to not only consider the loss landscape's properties to obtain higher accuracy, but also real accelerating ratios. To take into account factors of the loss landscape, a new concept \emph{recoverability} is proposed in this paper, which is computed efficiently with only a few training images. Our recoverability metric is surprisingly consistent with the accuracy of the finetuned network. Combined with the acceleration ratio, our final pruning score outperforms previous ones significantly.

Finally, we propose \PractiseNoSpace, namely \underline{Pr}actical network \underline{ac}celeration with \underline{ti}ny \underline{se}ts of images, to effectively accelerate a network with scarce data. \Practise significantly outperforms previous few-shot pruning methods. For $22.1\%$ latency reduction, \Practise surpasses the previous state-of-the-art (SOTA) method on average by \emph{7.0\%} (percentage points, not relative improvement) Top-1 accuracy on ImageNet-1k. It enjoys high generalization ability, which can be used on synthesized/out-of-domain images. Our contributions are:
\begin{itemize}
\item \textbf{Hypothesis.} We propose the finetune convexity hypothesis which unveils the mystery why \emph{few-shot} finetuning a pruned model does not suffer from the overfitting problem. We point out that convexity is tightly correlated with the effectiveness of finetuning. 

\item \textbf{Theory.} We establish a theory for feature mimicking, proving that learnable parameters achieve lower variance via feature mimicking compared to vanilla finetuning. It aligns very well with empirical results.

\item \textbf{Algorithm.} Although dropping blocks is previously known, we are \emph{the first to reveal its great potential in few-shot compression}, which has higher convexity than filter pruning. By considering properties of the optimization landscape (such as the convexity and initial accuracy), a new concept \emph{recoverability} is proposed to measure the difficulty of recovering each block, and in determining the priority to drop blocks. Then, we propose \PractiseNoSpace, an algorithm for accelerating networks with tiny sets of images. With 22.1\% latency reduction, \Practise outperforms previous SOTAs by 7.0\% Top-1 accuracy on ImageNet.
\end{itemize}

A preliminary version of the \Practise algorithm has appeared as a conference publication \cite{wang2023practical}.

\section{Related Work} 

\textbf{Filter-level pruning} accelerates networks by removing filters in convolutional layers. Different criteria for choosing filters have been proposed~\cite{he2019filter,molchanov2019importance,L1_norm,slimming,thinet,CURL}, along with different training strategies~\cite{liu2018rethinking,ResRep,li2022revisiting,wang2023EVC}. However, most of these methods rely on the whole training data to train the network. When facing data privacy issues, these filter-level pruning methods suffer from poor latency-accuracy performance with tiny training sets~\cite{FSKD,CD}. In this paper, we argue that the drawbacks of filter pruning are its flat optimization landscape and low acceleration ratio. It requires pruning lots of parameters and FLOPs to reduce latency. That results in a large capacity gap between the pruned and the original networks, and it is challenging to recover the pruned network's accuracy on only a tiny training set. Instead, we advocate block-level pruning for few-shot compression. Dropping blocks enjoys both a more convex optimization and higher acceleration ratio.

\begin{figure*}
	\centering
   \includegraphics[width=0.8\linewidth]{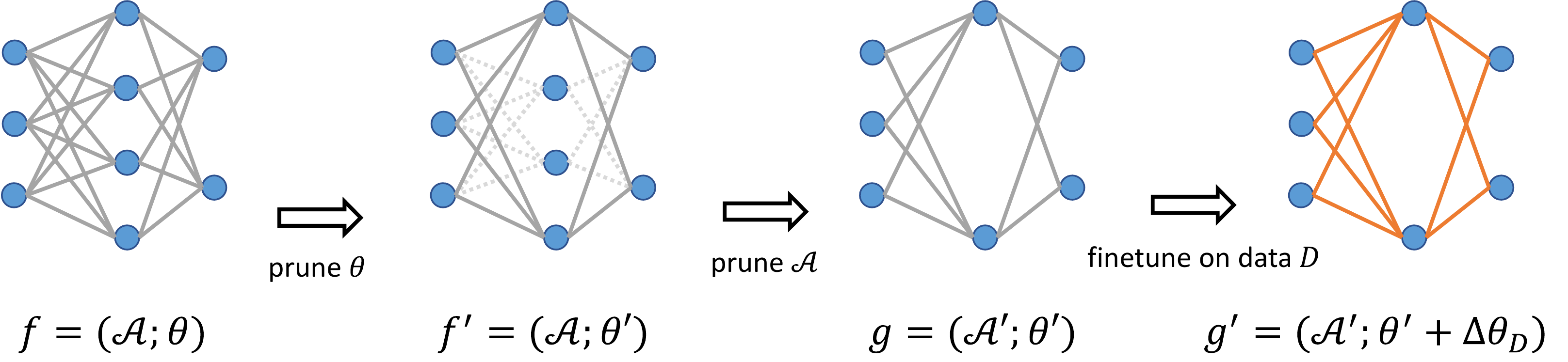}
  \caption{ The pipeline for traditional few-shot compression. }
  \label{fig:vanilla_FSP} 
\end{figure*}

\textbf{Block-level pruning} removes several complete blocks (e.g., residual blocks) in a network. Some work have been proposed for the whole training data case. DBP~\cite{DBP} proposes using linear probing to evaluate the accuracy of each block's features, and dropping blocks with low accuracy. $\epsilon$-ResNet adds a sparsity-promoting function to discard the block if all responses of this block are less than a threshold $\epsilon$. Both DBP and $\epsilon$-ResNet require a large dataset for training and testing. CURL~\cite{CURL} uses a proxy dataset to evaluate the KL-divergence change before/after block removal. However, it neglects the finetuning process. In fact, the accuracy of the pruned network \emph{after finetuning} is vital, but there are no existing criteria to approximate it well. In this paper, we propose a new \emph{recoverability} to evaluate if the network pruned by dropping blocks can recover the accuracy well.

\textbf{Few-shot or data-free knowledge distillation} aims at training a student network by a pretrained teacher with limited or even no original training data. Few-shot KD (FSKD)~\cite{FSKD} inserts $1\times 1$ conv. after the pruned conv. layer and trains each layer by making the pruned network's feature maps mimic the original network's. Layer-wise training can obtain more supervision from the teacher, but results in error accumulation. CD~\cite{CD} proposes cross distillation to reduce the accumulated errors. MiR~\cite{MiR} proposes a mimicking then replacing framework to optimize the pruned network holistically. When not even a single original training sample is available, data free knowledge distillation (DFKD)~\cite{DI} synthesizes alternatives of the original training data. Subsequent DFKD methods accelerate the synthesis process~\cite{fang2022up} and enhance the performance by multi-teacher~\cite{MixMix}. Pruning and quantization~\cite{liu2021zero,zhang2021diversifying,cai2020zeroq} are two main applications of DFKD. However, most FSKD and DFKD methods adopt filter-level pruning and result in poor latency-accuracy performance. Our \Practise works robustly on synthesized/out-of-domain images and improves the accuracy in the data-free scenario by a large margin. 

\section{Finetune Convexity Hypothesis}
\label{sec:loss-convex}

For few-shot compression, several effective algorithms~\cite{FSKD,CD,MiR} have been proposed. However, they are designed empirically and lack theoretical analysis. The key question is: why is it possible to finetune the network with only a few training samples without serious overfitting? In this section, we study the loss landscapes w.r.t. the parameters. Surprisingly, we find that finetuning with few samples appears to behave like a convex problem, which alleviates the overfitting issue. We observed that for different compression schemes, the convexity of their optimization landscapes is different. Pruning blocks enjoys a higher convexity, hence it requires less training data than filter pruning.

We formulate traditional few-shot compression into three steps as illustrated in Fig.~\ref{fig:vanilla_FSP}. First, some parameters are pruned by setting them to zero. For dropping a block, we prune all parameters inside the residual connection. For pruning filters, the pruned parameters are set to zero. Second, the network architecture is pruned guided by these zero parameters, while the network functionality is not changed. Third, the compressed network is finetuned on data $D$.

\begin{figure*}
  \centering
 \subcaptionbox{\label{fig:Loss}}{\includegraphics[width=0.325\linewidth]{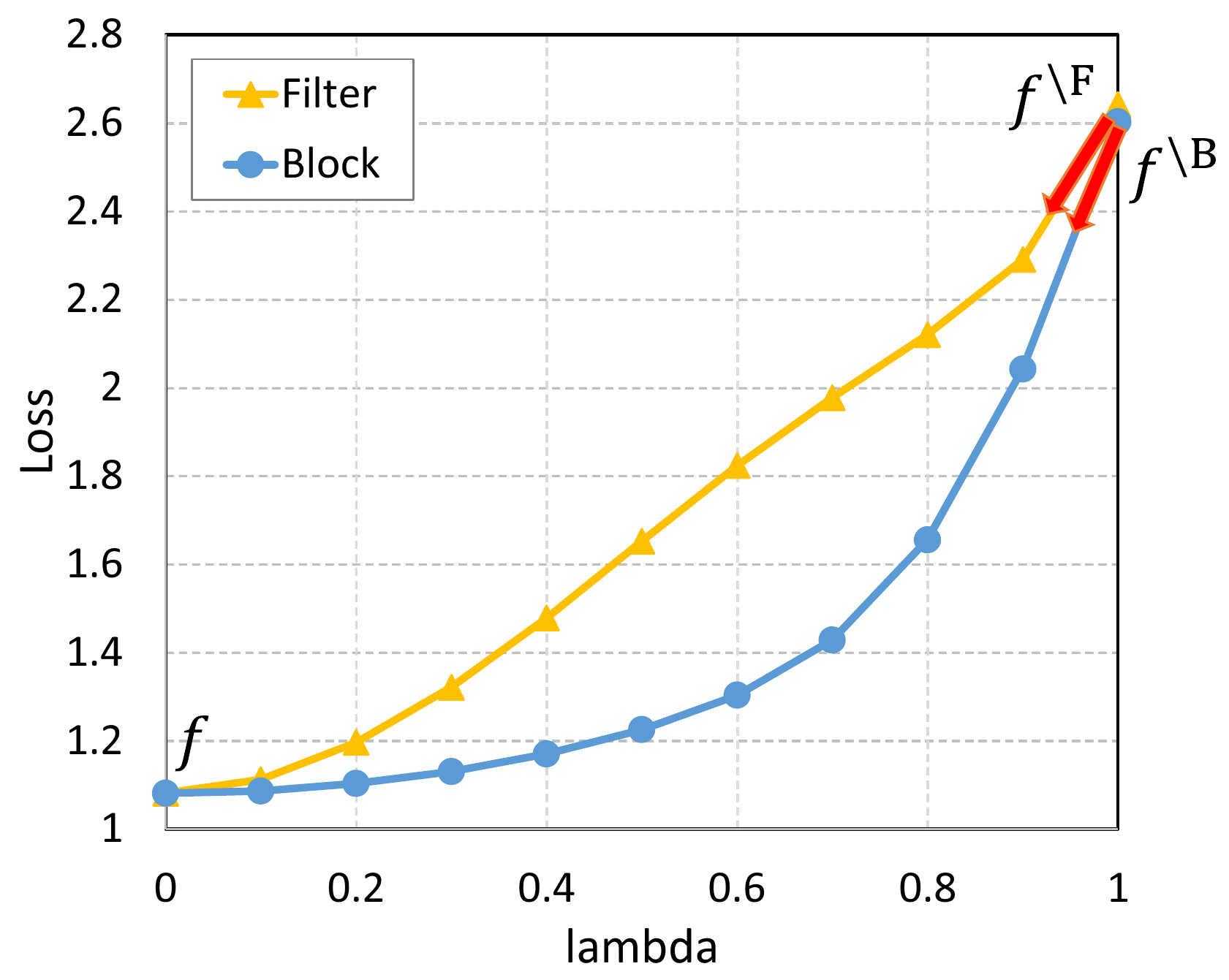}}
  \,
  \subcaptionbox{\label{fig:loss_pft}}{\includegraphics[width=0.325\linewidth]{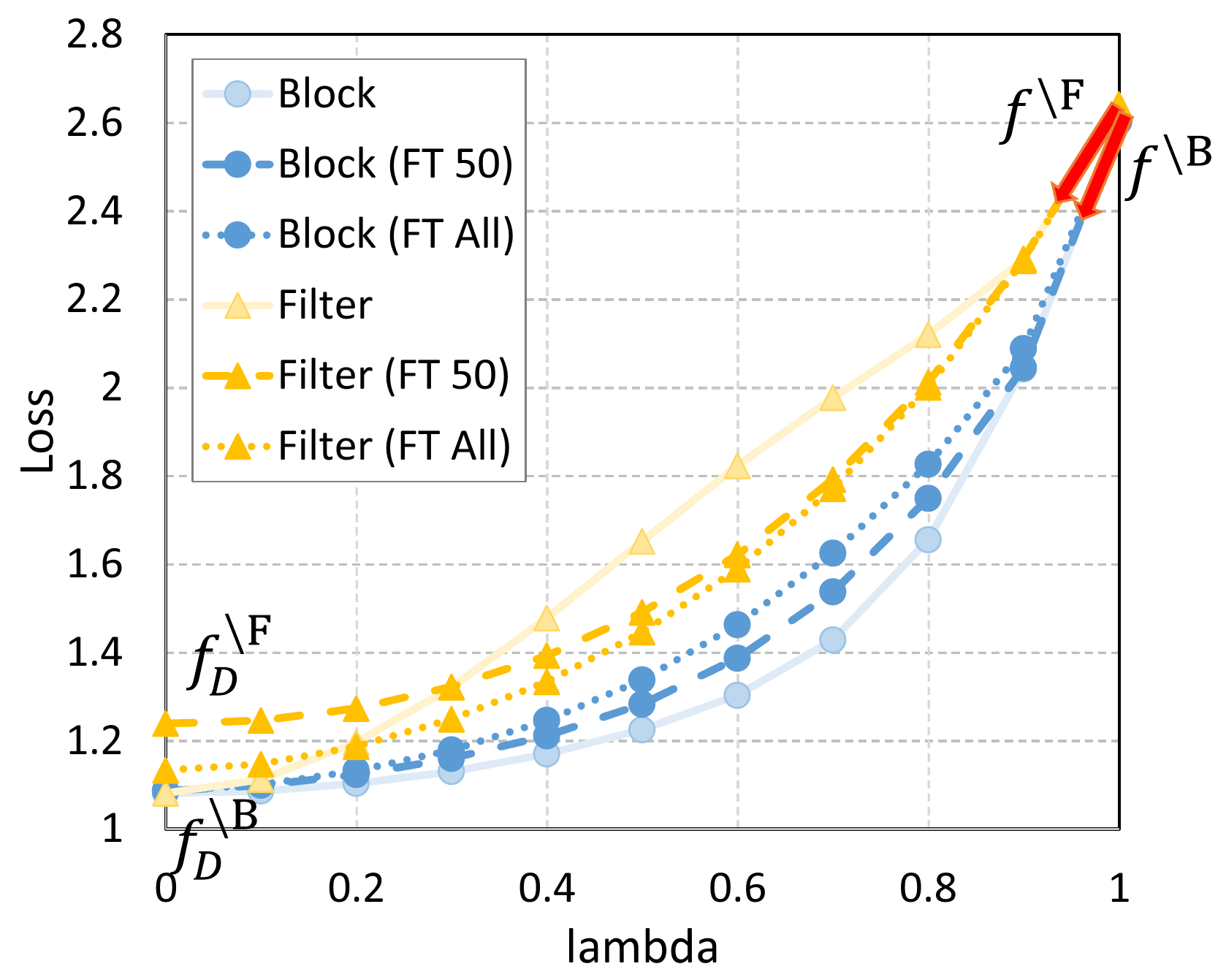}}
 \,
 \subcaptionbox{\label{fig:prune_param_land}}{
  \includegraphics[width=0.23\linewidth]{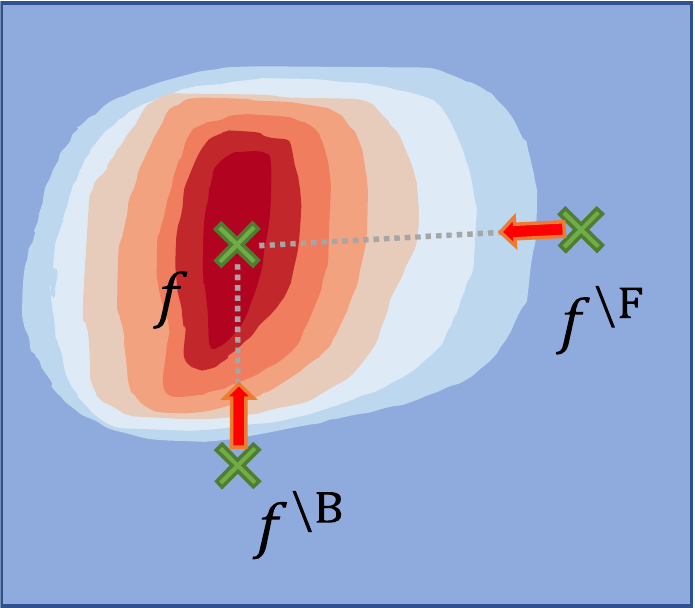}
  \vspace{20pt}
  }
\caption{(\protect\subref{fig:Loss}) presents the loss of interpolated models between the raw model $f$ ($\lambda=0$) and the pruned model $f'$ ($\lambda=1$). Then we finetune the pruned model $f'$ on datasets $D$ containing 50 (``FT 50'') or all (``FT All'') training images, and obtain the finetuned models $f'_D$. (\protect\subref{fig:loss_pft}) presents the loss of interpolated models between these two models. (\protect\subref{fig:prune_param_land}) illustrates the loss landscape qualitatively. Finetuning a pruned model enjoys a convex loss landscape w.r.t. model parameters. Compared to pruning filters, the loss landscape of pruning blocks is more convex. }
\label{fig:Raw2P} 
\end{figure*}

\begin{figure*}
  \centering
  \subcaptionbox{\label{fig:BFT_loss}}{\includegraphics[width=0.31\linewidth]{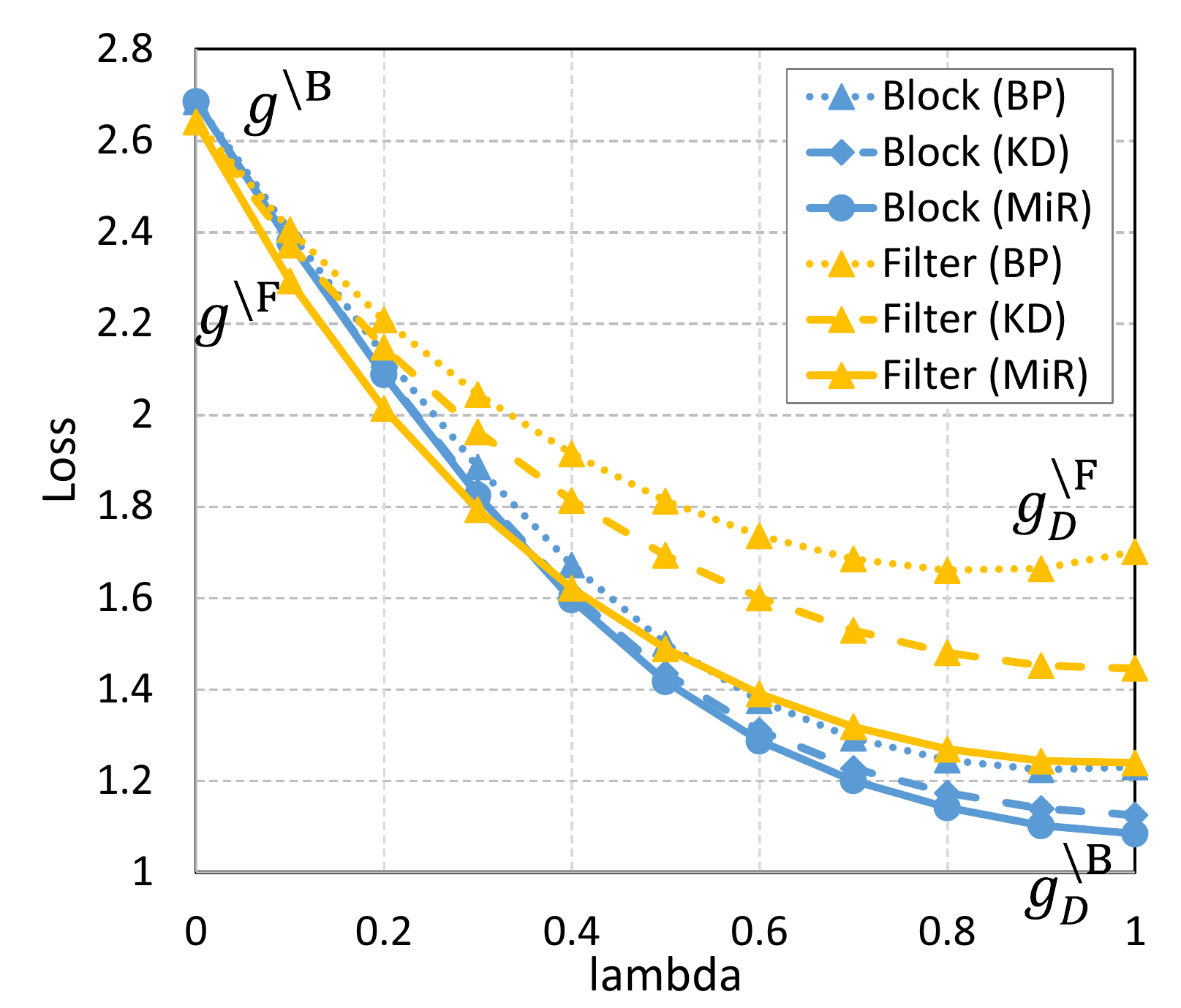}}
 \subcaptionbox{\label{fig:AB_loss}}{\includegraphics[width=0.4\linewidth]{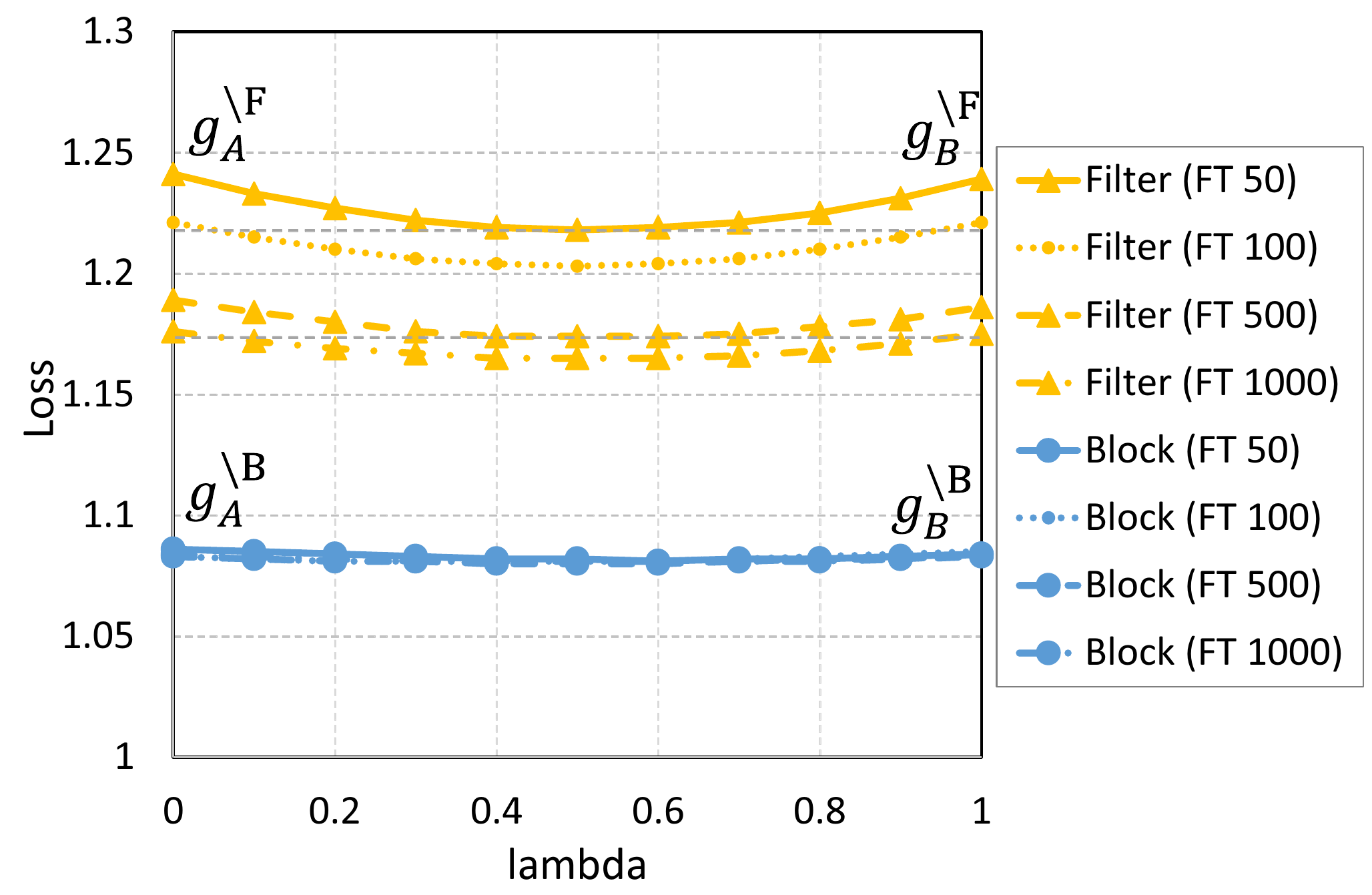}}
 \quad
 \subcaptionbox{\label{fig:FT_land}}{
  \includegraphics[width=0.25\linewidth]{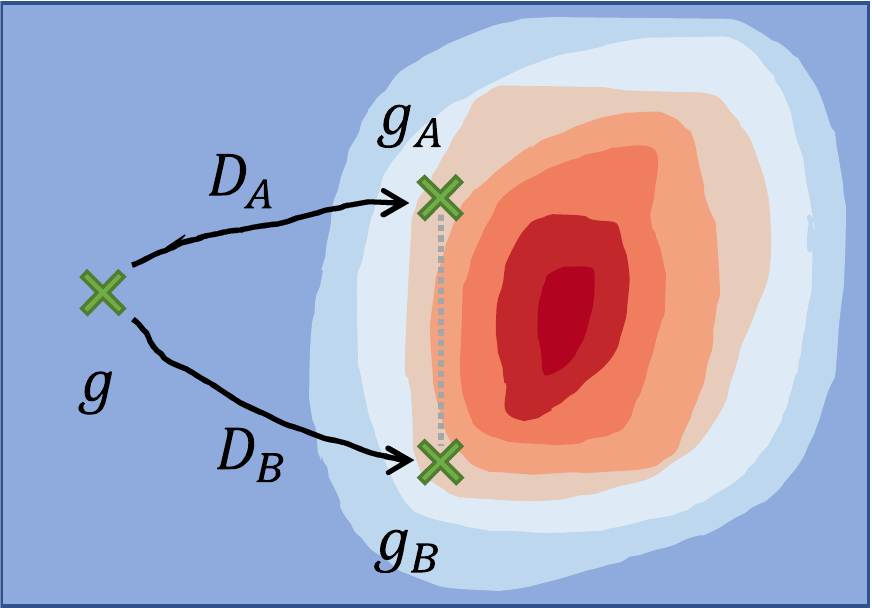}
  \vspace{20pt}
  }
\caption{(\protect\subref{fig:BFT_loss}) presents the loss of interpolated models with different finetuning methods. All models are finetuned by 50 samples. For (\protect\subref{fig:AB_loss}), we finetune models on two datasets $A$ and $B$ with the same amount, respectively. Then the loss of interpolations is evaluated. (\protect\subref{fig:FT_land}) illustrates the loss landscape of finetuning pruned models. }
\label{fig:AB} 
\end{figure*}

Following~\cite{draxler2018essentially,frankle2020linear,git_rebasin}, we measure the convexity via interpolating two models. Considering the first step $(\mathcal{A};\theta)\rightarrow (\mathcal{A};\theta')$ in Fig.~\ref{fig:vanilla_FSP}, we simply set some parameters to zero while keeping the architecture unchanged. Hence, it is able to merge the raw/pruned models by linear interpolation. Given $\lambda\in[0, 1]$, the interpolated model is 
\begin{align}
f_{\lambda} &=\lambda f' + (1-\lambda)f \\
\Longleftrightarrow (\mathcal{A};\theta_{\lambda}) &= (\mathcal{A};\lambda\theta'+(1-\lambda)\theta)\,.
\end{align}
We take ResNet-34 as the raw model ($f$), and prune it by dropping filters ($f^{\backslash F}$) and blocks ($f^{\backslash B}$), respectively. Fig.~\ref{fig:Loss} shows the test loss of interpolated models. The loss curve roughly describes the property of the loss landscape. Surprisingly, these curves are smooth and monotonic. More importantly, they are roughly convex. Pruning blocks achieves a more convex curve compared to pruning filters. Fig.~\ref{fig:prune_param_land} presents this observation qualitatively. Compare to the path $f\rightarrow f^{\backslash F}$, the path $f\rightarrow f^{\backslash B}$ is more convex. With these observations, we propose the following conjecture.

\begin{conjecture}
  \label{conj:convexity_of_pruning}
  By linearly interpolating between the raw and the pruned model, the convexity of the loss landscape indicates the difficulty of finetuning, which reflects the amount of training data required, the speed of convergence, etc.
\end{conjecture}

To check this conjecture, we finetune both $f^{\backslash B}$ and $f^{\backslash F}$ on a training set $D$, and obtain the finetuned models $f^{\backslash B}_D$ and $f^{\backslash F}_D$, respectively. Then we evaluate interpolations between pruned and finetuned models. Fig.~\ref{fig:loss_pft} shows results. First, with a more convex loss landscape, $f^{\backslash B}_{D}$ achieves a lower loss than $f^{\backslash F}_{D}$ in all cases. Intuitively, a more convex loss landscape means it is easier for SGD to optimize models rapidly. Second, for $\lambda\in[0.9, 1]$, the dash lines overlap on the solid lines, that is because the interpolation equation $\theta_{\lambda}=\lambda\theta^{\backslash B}+(1-\lambda)\theta$ can be considered as the SGD updating formula, where $\theta$ is the optimal gradient direction and $1-\lambda$ is the small learning rate. This further demonstrates the validity of Conj.~\ref{conj:convexity_of_pruning} based on the close relationship between linear interpolation and SGD finetuning.

Then we consider the remaining steps in few-shot compression, where the architecture will be pruned. Assume $f^{\backslash B}$ and $f^{\backslash F}$ become $g^{\backslash B}$ and $g^{\backslash F}$ by pruning architecture, respectively. We finetune these pruned models on a tiny data $D$, and obtain $g^{\backslash B}_D$ and $g^{\backslash F}_D$. Fig.~\ref{fig:BFT_loss} shows the results of linear interpolation. All curves of different finetuning methods are convex, and pruning blocks outperforms pruning filters for lower loss significantly. Based on both Fig.~\ref{fig:loss_pft} and \ref{fig:BFT_loss}, we propose the following conjecture.
\begin{conjecture}
  \label{conj:convexity_of_finetune}
  \textbf{[Finetune convexity hypothesis (weak)]:} The loss landscape of finetuning a pruned model is convex w.r.t. the parameters. That means
  \begin{equation}
    \mathcal{L}(g'_{\overset{\lambda}{\leftrightarrow} D}) \leq \lambda \mathcal{L}(g) + (1 - \lambda) \mathcal{L}(g'_D)\,,
  \end{equation}
  where $g'_{\overset{\lambda}{\leftrightarrow} D}$ is an interpolated model:
  \begin{align}
     g'_{\overset{\lambda}{\leftrightarrow} D} &= (\mathcal{A}';\theta'+(1 - \lambda)\Delta\theta_D)\,, \\
     g &= (\mathcal{A}';\theta')\,, \\
     g'_D &= (\mathcal{A}';\theta'+\Delta\theta_D)\,.
  \end{align}
\end{conjecture}

\begin{figure}
	\centering
  \subcaptionbox{\label{fig:TFS_loss}}{\includegraphics[width=0.45\linewidth]{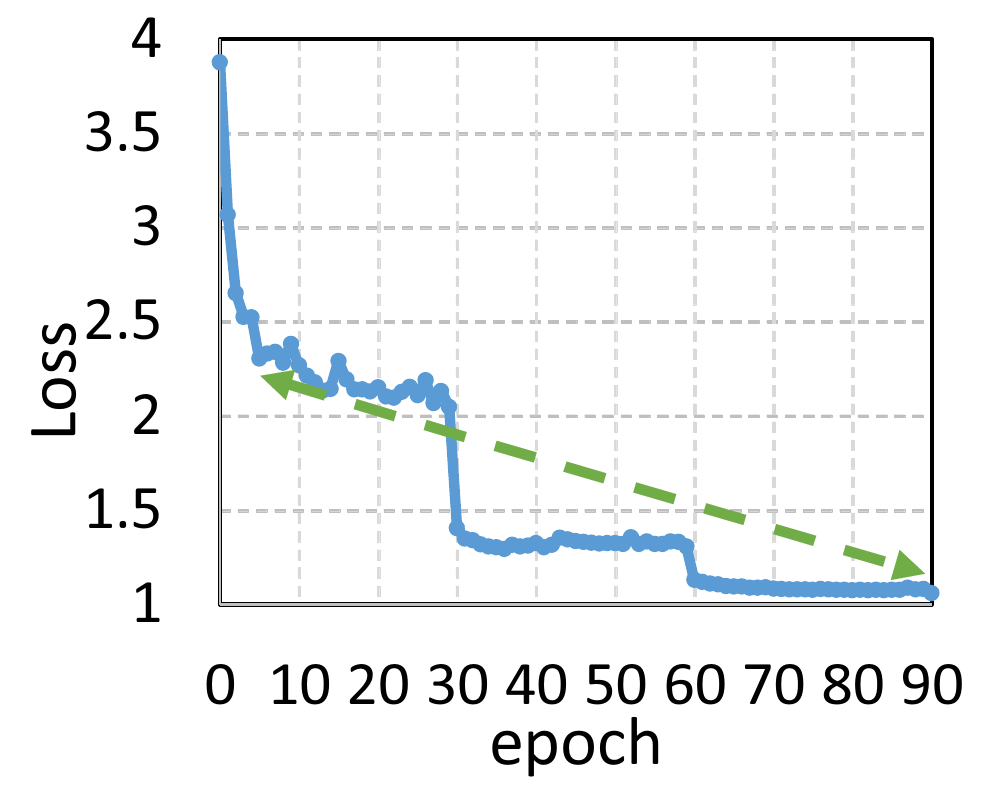}}
	\,
   \subcaptionbox{\label{fig:TFS_merge}}{\includegraphics[width=0.45\linewidth]{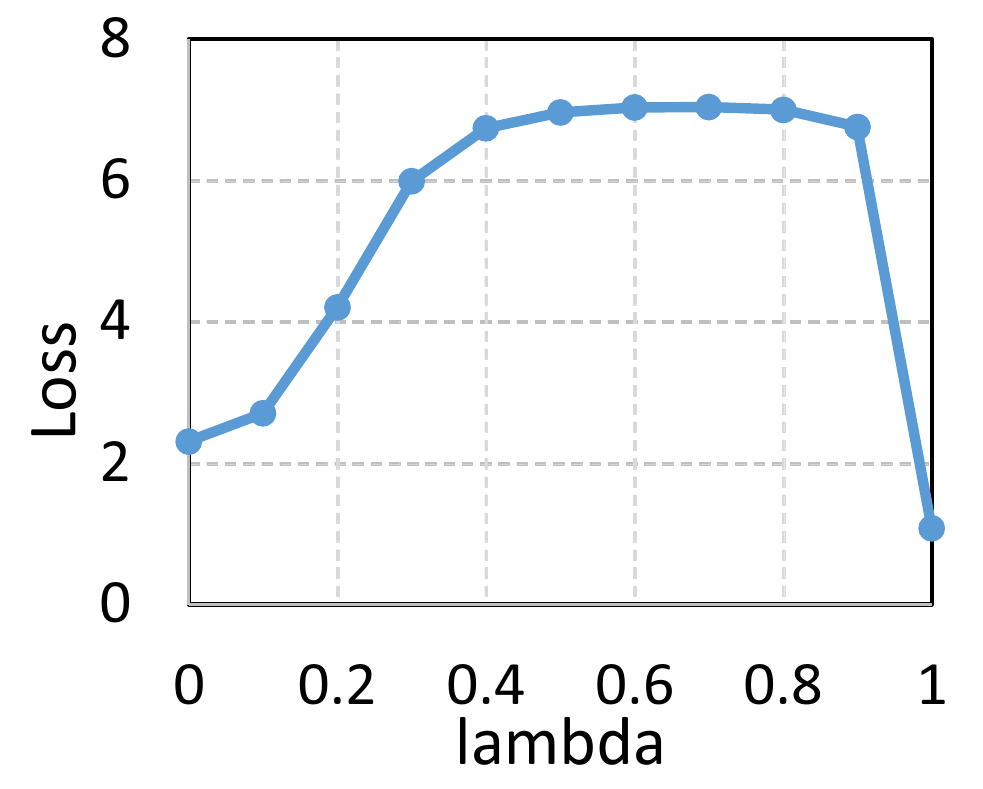}}
  \caption{(\protect\subref{fig:TFS_loss}) shows the testing loss curve of pruned model $g^{\backslash B}$ on the whole ImageNet-1k training set. (\protect\subref{fig:TFS_merge}) presents the loss of interpolated models between the 5th epoch model ($\lambda=0$) and the last epoch model ($\lambda=1$), i.e., those indicated by green arrows in (\protect\subref{fig:TFS_loss}).}
  \label{fig:TFS} 
\end{figure}

This conjecture answers the question of why it is able to finetune the whole network with only a few training samples. With the convexity property, optimization hardly suffers from overfitting even if only scarce training data is available. Note that \emph{this conclusion does not hold for training from scratch with the whole training data}. As introduced in previous work~\cite{draxler2018essentially,frankle2020linear,git_rebasin}, a loss barrier exists on linear interpolation between the raw and finetuned models. We also evaluate it by training a pruned network from scratch. Fig.~\ref{fig:TFS} shows results. Training from scratch suffers from a complex optimization landscape. Linearly interpolating cannot achieve a convex loss curve.

Conj.~\ref{conj:convexity_of_finetune} only considers the connection between a model $g$ and its finetuned version $g'$. We further finetune the pruned model $g$ on different data sets $A$ and $B$ to study the relationship between different finetuned versions. Fig.~\ref{fig:AB_loss} presents results of linear interpolations between $g'_A$ and $g'_B$. With different tiny sets, the finetuned models are in the same basin, and there is no loss barrier on interpolation. Fig.~\ref{fig:FT_land} presents this observation qualitatively. Hence, we have the following strong conjecture:

\begin{conjecture}
  \label{conj:Loss_convex}
  \textbf{[Finetune convexity hypothesis (strong)]:} With few-shot training datasets $A$ and $B$, the finetuning solutions lie in the same optimization basin and the functional $\mathcal{L}$ is convex w.r.t. the parameters $\theta$. That means 
  \begin{equation}
    \mathcal{L}(g'_{A\overset{\lambda}{\leftrightarrow} B}) \leq \lambda \mathcal{L}(g'_A) + (1 - \lambda) \mathcal{L}(g'_B)\,,
  \end{equation}
  where
  \begin{align}
     g'_{A\overset{\lambda}{\leftrightarrow} B} &= (\mathcal{A};\theta'+\lambda\Delta\theta_A+(1 - \lambda)\Delta\theta_B)\,, \\
     g'_A &= (\mathcal{A};\theta'+\Delta\theta_A)\,, \\
     g'_B &= (\mathcal{A};\theta'+\Delta\theta_B)\,.
\end{align}
We define the loss leakage as 
  \begin{align}
    &\mathcal{L}_{leak}(g'_{A\leftrightarrow B}) = \max_{\lambda} \lambda \mathcal{L}(g'_A) + (1 - \lambda) \mathcal{L}(g'_B) - \mathcal{L}(g'_{\lambda})\,.
 \end{align}
If the training data is increased from $A$ to $A \cup B$, the loss reduction is proportional to the loss leakage. That is
 \begin{align}
  \mathcal{L}(g'_A) - \mathcal{L}(g'_{A \cup B}) \propto \mathcal{L}_{leak}(g'_{A\leftrightarrow B}) \,.
 \end{align}
\end{conjecture}

Our loss leakage indicates the pruned network's thirst for training data. With a large loss leakage, the pruned network requires more data to finetune. As illustrated in Fig.~\ref{fig:AB_loss}, compared to pruning blocks, pruning filters needs more data to reduce the loss. In particular, for A and B containing 50 images each, the linear interpolation model ($\lambda=0.5$) achieves a similar loss with the model finetuned on 100 images. In other words, averaging models finetuned on each set separately is on-par with finetuning the model on the union set. This observation supports our Conj.~\ref{conj:Loss_convex} and further supports our Conj.~\ref{conj:convexity_of_finetune}. Note that Conj.~\ref{conj:Loss_convex} can be considered as an extension of Conj.~\ref{conj:convexity_of_finetune} by setting an empty dataset $A$. 

In summary, we have unraveled the mysteries proposed at the beginning of this section. Thanks to the convexity of the optimization landscape, finetuning with few training samples is possible without serious overfitting. The loss landscape of pruning blocks is more convex compared to pruning filters. Hence, it requires less training data and enjoys a faster convergence. Next, we will propose a theory for different finetuning methods based on our convexity hypothesis.

\section{Theory on Finetuning}
\label{sec:theory}

As presented in Fig.~\ref{fig:BFT_loss}, different finetuning methods lead to different final accuracies in practice. The naive method (``BP'') simply uses the cross-entropy loss, which results in inferior accuracy as previous work pointed out~\cite{FSKD,CD}. Knowledge distillation~\cite{KD} (KD), especially feature-level distillation~\cite{fitnet,LSHKD}, can improve accuracy. MiR~\cite{MiR} proposes to use the features before the global average pooling for the pruned model to mimic and achieved state-of-the-art performance with few training samples. Note that all these algorithms were designed empirically. There is no theory to explain why the feature mimicking method MiR outperforms other methods. In this section, we propose one that explains the behaviors of existing few-shot distillation approaches, and shows that the loss function of MiR is optimal among them. Furthermore, we show that feature mimicking achieves a more stable model compared to optimizing cross-entropy. 

\begin{figure}
	\centering
   \subcaptionbox{\label{fig:vanilla}\quad}{\includegraphics[width=0.25\linewidth]{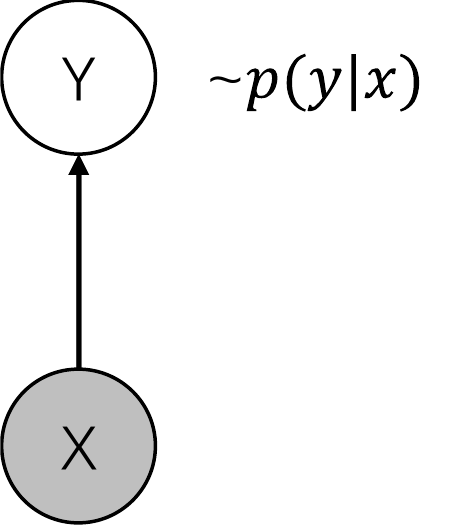}}
	\qquad
   \subcaptionbox{\label{fig:FM}\quad}{\includegraphics[width=0.25\linewidth]{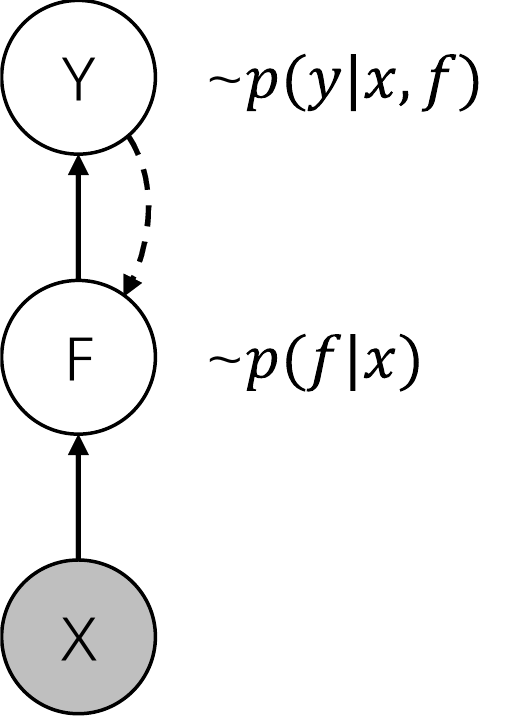}}
  \caption{(\protect\subref{fig:vanilla}) and (\protect\subref{fig:FM}) present the vanilla discriminative model and feature mimicking model, respectively.}
  \label{fig:pro_model} 
  \end{figure}

We formulate traditional classification as the discriminative model in Fig.~\ref{fig:vanilla}. Given the input $x$, this model estimates the conditional probability distribution $p(y|x)$. With KD, $p(y|x)$ can be viewed as the teacher's distribution, while the student distribution is $q_\theta (y|x)$, where $\theta$ denotes parameters. KL-divergence is adopted to measure the distance between these two distributions: 
\begin{equation}
   \label{eq:vanilla}
   KL\left[p(y|x)||q_\theta(y|x)\right]\,.
\end{equation}
For simplicity, we only consider the single variate case, but it is natural to extend the analyses to multivariate.

More information about intermediate features is introduced in feature mimicking, shown as Fig.~\ref{fig:FM}. We know the feature distribution $p(f|x)$ in addition to the teacher's prediction $p(y|x, f)$. Now, the target is the joint distribution of both $f$ and $y$:
\begin{equation}
   \label{eq:FM}
   KL\left[p(y, f|x)||q_\theta(y, f|x)\right]\,,
\end{equation}
where $p$ and $q$ denote the teacher and student, respectively. Intuitively, once we successfully optimize this target, we can assume $q_\theta(y, f|x)$ approaches $p(y, f|x)$, denoted by $q_\theta(y,f|x)\rightarrow p(y,f|x)$. Therefore, 
\begin{equation}
  q_\theta(y|x)=\int q_\theta(y,f|x)df\rightarrow\int p(y,f|x)df=p(y|x)\,,
\end{equation}
which means we also optimize Eq.~\ref{eq:vanilla} successfully and obtain a discriminative model for classification. The following claim establishes a strict connection between Eq.~\ref{eq:vanilla} and Eq.~\ref{eq:FM}. Its proof is in Appendix~\ref{app:proof:claim1}. 

\begin{claim}
   \label{claim:upper-bound}
   Eq.~\ref{eq:vanilla} is upper bounded by Eq.~\ref{eq:FM}, that is,
   \begin{equation}
    KL\left[p(y|x)||q_\theta(y|x)\right] \leq KL\left[p(y, f|x)||q_\theta(y, f|x)\right]\,.
   \end{equation}
\end{claim}

Claim~\ref{claim:upper-bound} tells us optimizing Eq.~\ref{eq:FM} is reasonable for obtaining a classifier. In the rest of this section, we will show that optimizing Eq.~\ref{eq:FM} is easier than Eq.~\ref{eq:vanilla}, and it results in a more stable model. First, we introduce how to optimize Eq.~\ref{eq:FM} by the following claim, whose proof is in Appendix~\ref{app:proof:claim2}. 

\begin{claim}
  \label{claim:loss-decouple}
  Eq.~\ref{eq:FM} can be decoupled into a classification loss and a feature mimicking loss, in the form
  \begin{align}
  & KL\left[p(y, f|x)||q_\theta(y, f|x)\right] = C \,+ \notag\\
  &\underset{f\sim p(f|x)}{\mathbb{E}} \left\{\underbrace{- \ln q_\theta(f|x)}_{\text{Feature Mimicking}} + \underbrace{KL\left[p(y|f,x) || q_\theta(y|f,x)\right]}_{\text{Classification}}\right\}\,,
  \end{align}
  where $C$ is a constant. Assume $q_\theta(f|x)$ is a Gaussian with mean $\mu_\theta(x)$ and standard deviation $\frac{1}{\sqrt{2\beta}}$, then
  \begin{align}
    \label{eq:MSE_loss}
    &\,\mathbb{E}_{f\sim p(f|x)} [- \ln q_\theta(f|x)] \notag \\
    = &\, \mathbb{E}_{f\sim p(f|x)} \left[\beta\left(f-\mu_\theta(x)\right)^2 \right]+C'\,.
  \end{align}
\end{claim}

That is, using the L2 loss for the feature mimicking component is optimal under certain reasonable assumptions. In practice, this loss is often adopted as the feature-based distillation loss. $\beta$ is then a hyperparameter to balance feature mimicking and classification, and $\mu(x)$ is our student backbone network which generates features given input $x$. 

\begin{figure}
	\centering
  \subcaptionbox{\label{fig:KD} vanilla KD}{\includegraphics[width=0.32\linewidth]{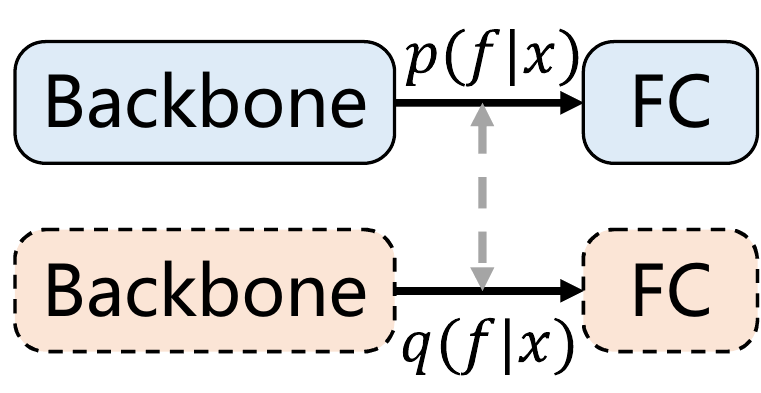}}
   \subcaptionbox{\label{fig:CD} cross distillation}{\includegraphics[width=0.32\linewidth]{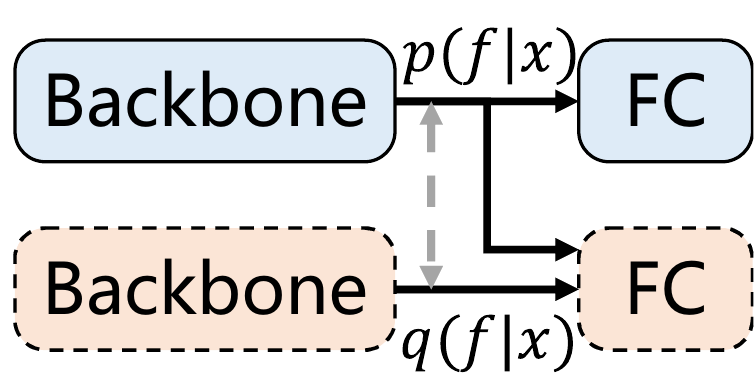}}
   \subcaptionbox{\label{fig:MiR} MiR}{\includegraphics[width=0.32\linewidth]{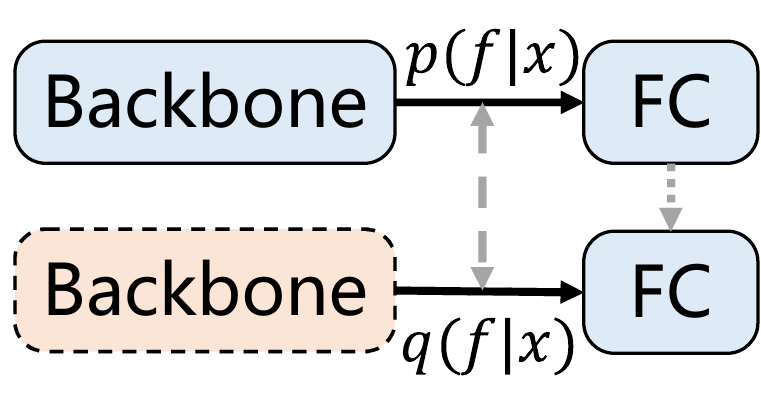}}
  \caption{Different ways to optimize the classifier (``FC'') in KD. Blue modules with solid boundaries denote teacher modules, while orange modules with dash boundaries denote student modules. }
  \label{fig:class} 
\end{figure}

For classification loss $\mathbb{E}_{f\sim p(f|x)} KL\left[p(y|f,x) || q_\theta(y|f,x)\right]$, we notice $f$, which is generated from the teacher $p(f|x)$, appears in $q_\theta(y|f,x)$. Hence, the student can utilize the teacher's features during training. We summarize three ways to implement this loss, as illustrated in Fig.~\ref{fig:class}.

\begin{enumerate}
	\item To ignore the teacher's $f$, and directly model $y$ according to the input $x$ (as in Fig.~\ref{fig:KD}). This is the traditional manner widely used in previous work~\cite{KD,LSHKD,fitnet,FSKD}. Without the help of the teacher's feature, the model achieves inferior accuracy. 
	\item Second, cross distillation (as in Fig.~\ref{fig:CD}). The student classifier utilizes the teacher features to reduce the classification loss. Cross distillation outperforms vanilla distillation. This approach was used in CD~\cite{CD} and grafting~\cite{graft} methods.
	\item Directly use the teacher's classifier (as in Fig.~\ref{fig:MiR}). Because we already know $p(y|f)$ (which is the teacher's classifier or ``FC''), we can directly use it for $q_\theta(y|f)$. This is the optimal solution, making $\mathbb{E}_{f\sim p(f|x)} KL\left[p(y|f,x) || q_\theta(y|f,x)\right]=0$. This approach was used in MiR~\cite{MiR}, and also used in this paper as the finetuning method.
\end{enumerate}

\subsection{Variance on model parameters}

Next, we analyze why Eq.~\ref{eq:FM} is better than Eq.~\ref{eq:vanilla}. Our main result is: feature mimicking (Eq.~\ref{eq:FM}) achieves a lower variance on the model's parameters. In other words, it requires less training data to obtain a better estimate of parameters. We start from the following claim (cf. Appendix~\ref{app:proof:claim3} for the proof).

\begin{claim}
  \label{claim:variance}
  Given $n$ training samples $x_1,x_2,\cdots,x_n$, assume the loss function for finetuning follows the canonical form
  \begin{equation}
    \mathcal{L} = -\ln q_\theta (f|x)\,.
  \end{equation}
  Then we approximately have 
  \begin{equation}
    \label{eq:var(theta)}
    \hat{\theta}\sim \mathcal{N}(\tilde{\theta}, \frac{1}{n\mathcal{I}(\tilde{\theta})})\,,
  \end{equation}
  where $\tilde{\theta}$ denotes the teacher parameters, and 
  \begin{equation}
    \mathcal{I}(\theta)=\mathbb{E}_{x\sim p(x)}\int \left(\frac{\partial q_\theta (f|x)}{\partial \theta}\right)^2\frac{1}{q_{\theta}(f|x)}df\,.
  \end{equation}
\end{claim}

Claim~\ref{claim:variance} establishes an estimate for the variance. According to Eq.~\ref{eq:var(theta)}, the variance of $\hat{\theta}$ is calculated by $\frac{1}{n\mathcal{I}(\theta_t)}$. In the following, we compute this term for feature mimicking and classification in Claims~\ref{claim:FM:variance} and \ref{claim:class:variance}, respectively. The proofs can be found in Appendix~\ref{app:proof:claim4} and \ref{app:proof:claim5}, respectively.

\begin{claim}
  \label{claim:FM:variance}
  If a network is finetuned by feature mimicking with the loss in Eq.~\ref{eq:MSE_loss}, and assume the last layer before the output features is linear (i.e., the convolution layer):
  \begin{equation}
    \mu_{w,b}(x)=w\cdot x + b\,,
  \end{equation}
  then by Claim~\ref{claim:variance}, we have
  \begin{align}
    \hat{w} &\sim \mathcal{N}(\tilde{w}, \frac{1}{2n\beta \mathbb{E}[x^2]})\,, \\
    \hat{b} &\sim \mathcal{N}(\tilde{b}, \frac{1}{2n\beta })\,. 
  \end{align}
\end{claim}

\begin{claim}
  \label{claim:class:variance}
  If a network is finetuned by optimizing the loss in Eq.~\ref{eq:vanilla}, and assume the last layer is
  \begin{equation}
    q_{\vw,\vb}(\vy|f)=\sigma(\vw\cdot f + \vb)\,,
  \end{equation}
  where $\sigma$ is the softmax operation 
  \begin{equation}
  \sigma(\vx)_i=\frac{\exp(x_i)}{\sum_{j=1}^{C}\exp(x_j)}\,,
  \end{equation}
  then by Claim~\ref{claim:variance}, for the $i$-th entry of $\vw$ and $\vb$, we have
  \begin{align}
    \hat{w}_i &\sim \mathcal{N}(\tilde{w}_i, \frac{1}{n\epsilon_i\mathbb{E}[f^2]})\,, \\
    \hat{b}_i &\sim \mathcal{N}(\tilde{b}_i, \frac{1}{n\epsilon_i})\,,
  \end{align}
  where $\epsilon_i$ is the $i$-th entry of $\boldsymbol{\epsilon}$ and
  \begin{equation}
    \boldsymbol{\epsilon} = q_{\vw,\vb}(\vy|f)[1-q_{\vw,\vb}(\vy|f)]\,.
  \end{equation}
\end{claim}

According to Claims~\ref{claim:FM:variance} and \ref{claim:class:variance}, the variances of the weight parameters for feature mimicking and classification are $\frac{1}{2n\beta\mathbb{E}[x^2]}$ and $\frac{1}{n\epsilon\mathbb{E}[f^2]}$, respectively. Because normalization layers exist in networks, we can assume both $\mathbb{E}[x^2]$ and $\mathbb{E}[f^2]$ are bounded by a constant. If a large-scale training set is available, both variances are small due to a large $n$.

But, if $n$ is small, the variance for classification (Claim~\ref{claim:class:variance}) is large, because $\epsilon\rightarrow 0$ empirically (about $10^{-3}$). For the feature mimicking, the typical value of $\beta$ is $\frac{1}{2}$, and hence the variance (Claim~\ref{claim:FM:variance}) is still small. That means in the few-shot scenario, \emph{feature mimicking is more stable than classification}.

Finally, our theory also explains traditional KD~\cite{KD}, which applies a temperature $T$ within the softmax operation:
\begin{equation}
  q_{\vw,\vb}(\vy|f)=\sigma(\frac{\vw\cdot f + \vb}{T})\,.
\end{equation}
If $T>1$, it will soften the prediction and make $\mathbf{\epsilon}$ larger. This explains why KD is very helpful in few-shot compression.

In short, our theory establishes that feature mimicking achieves lower variance on model parameters compared to optimizing the cross-entropy loss. This explains why feature mimicking is a superior finetuning method. 

\section{The \Practise Algorithm}
\label{sec:algorithm}

Now we move on to propose a new algorithm for few-shot compression aligned with our hypothesis and theory. First, we analyze the benefits of dropping blocks (e.g., residual blocks in ResNet). Compared with previous few-shot compression schemes, dropping blocks enjoys a more convex optimization landscape and a high acceleration ratio that achieves superior latency-accuracy performance with tiny training sets. Second, we propose a new recoverability criterion to choose which blocks to drop. Different from previous criteria, recoverability measures the hardness of finetuning a pruned network, which is not only closely and more directly related to the model's accuracy, but also is efficient to compute and suits the few-shot scenario well. Based on the recoverability, we propose \PractiseNoSpace, an algorithm for practical network acceleration with tiny training sets. \Practise does not need image labels, and it is even able to work in the data-free setting.

\subsection{The motivation to drop blocks}

\begin{figure}
	\centering
	\includegraphics[width=0.6\linewidth]{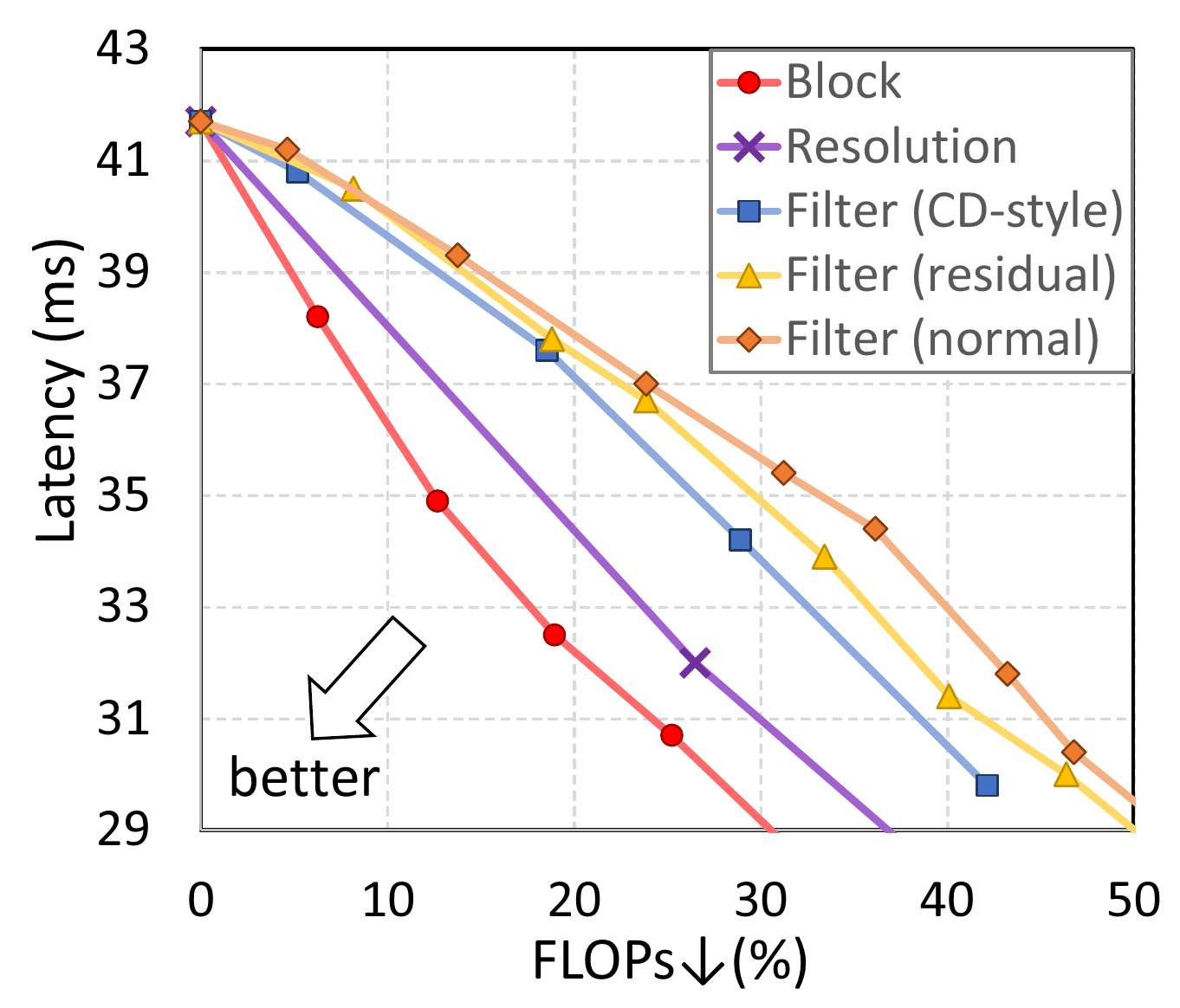}
	\caption{The relationship between latency reduction and FLOPs reduction. Dropping blocks enjoys the highest acceleration ratio compared with other compression schemes.}
	\label{fig:latency-flops}
\end{figure}

For accelerating neural networks in real-world applications, latency and accuracy are the two most important metrics. Lower latency (instead of FLOPs) means the model runs faster \emph{in reality}. To accelerate the model with tiny training sets, many compression schemes have been proposed. Fig.~\ref{fig:latency-flops} compares the acceleration ratios of them. FSKD~\cite{FSKD} prunes filters within residual blocks according to $L_1$ norm (`normal'). CD~\cite{CD} proposes pruning conv. layers only in shallow layers, and keeps deeper layers unchanged (`CD-style'). MiR~\cite{MiR} trims the residual connection (`residual'). All these pruning schemes suffer from inefficient acceleration ratios. As shown in Fig.~\ref{fig:latency-flops}, with about $30\%$ FLOPs reduction, compressed models achieve only 16.1\% latency reduction (41.7$\rightarrow$35 ms). Simply resizing the input image's resolution (namely `Resolution') achieves better acceleration. At last, dropping blocks outperforms all these methods. To achieve 35 ms latency, it only needs to reduce 12.7\% FLOPs, significantly less than the $30\%$ in pruning filters.

\begin{figure}
	\centering
	\includegraphics[width=0.86\linewidth]{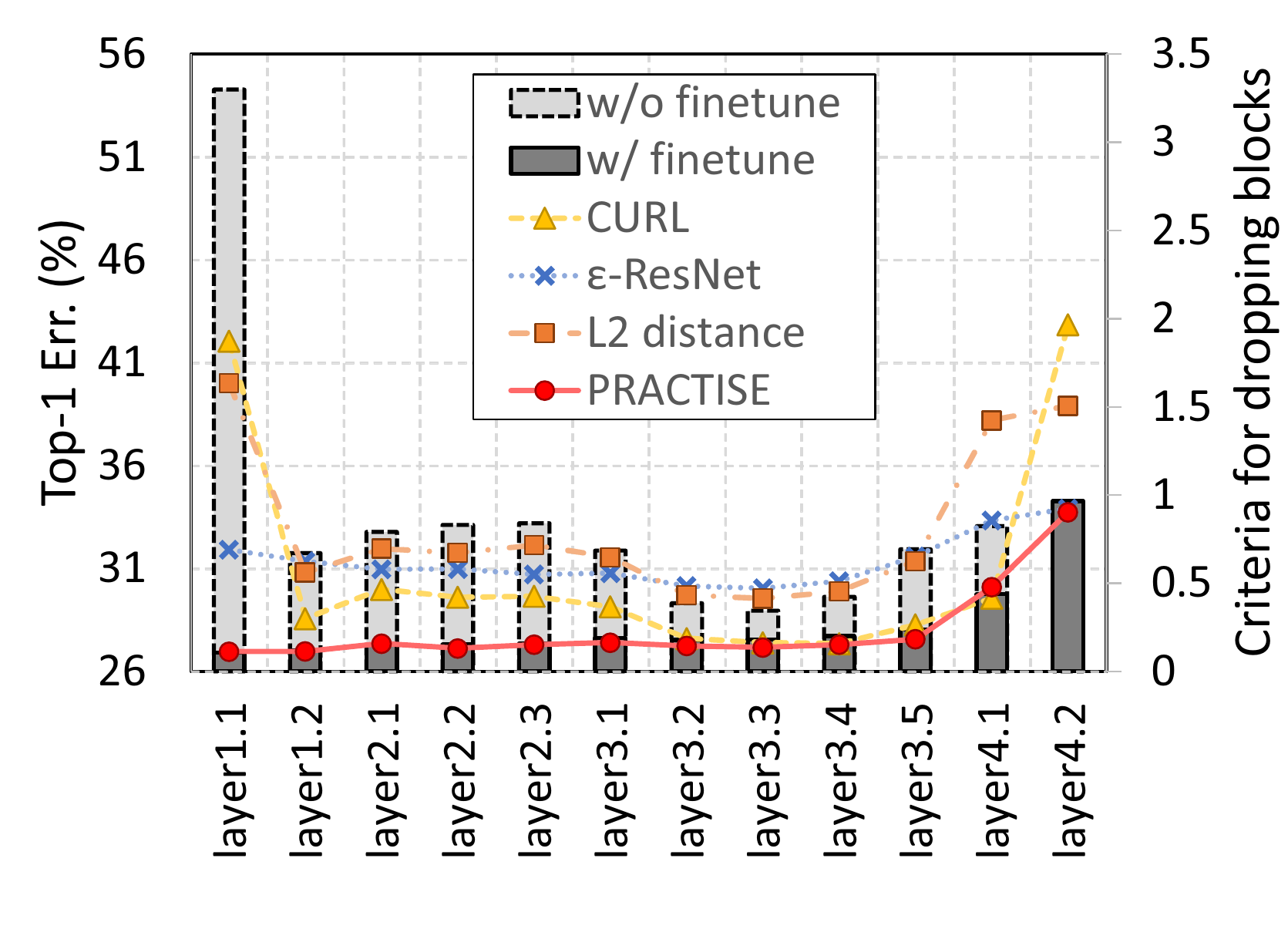}
	\caption{Illustration of the recoverability of each block and different criteria for dropping blocks. ResNet-34 has 12 blocks that can be dropped (`layer1.1' to `layer4.2'). We drop one block each time and evaluate the Top-1 error with or without finetuning. Scores of different criteria are computed for each block and presented in this figure as different curves, too. Note that the Top-1 error (\%) is evaluated on the ImageNet-1k validation set (50000 images), while finetuning and evaluating criteria take only 500 training images. Our \Practise predicts the finetuned network's error almost perfectly well.}
	\label{fig:metric}
\end{figure}

We also care about the accuracy of the finetuned model. As discussed in Sec.~\ref{sec:loss-convex}, pruning blocks enjoys a more convex optimization landscape compared to pruning filters, hence requires less training samples (cf. Fig.~\ref{fig:AB}). 

Obviously, dropping blocks is more effective for model acceleration than pruning filters, but it is neglected in few-shot compression. One possible reason is that pruning filters has achieved extraordinary performance with the \emph{whole} training dataset. However, when only a tiny training set is available, we should consider pruning efficiency more carefully. Existing work~\cite{FSKD,MiR} prune filters with large FLOPs reductions but suffer from a less convex optimization landscape, hence require more training data. On the other hand, dropping blocks enjoys a high acceleration ratio and keeps more parameters and capacity from the original model. With a more convex loss landscape, it requires less data for finetuning and achieves a superior latency-accuracy tradeoff with the \emph{tiny} training set (cf. Fig.~\ref{fig:top1-latency}). We have demonstrated that even a naive dropping block method has already outperformed most existing methods.

\begin{figure*}
	\centering
	\includegraphics[width=0.7\linewidth]{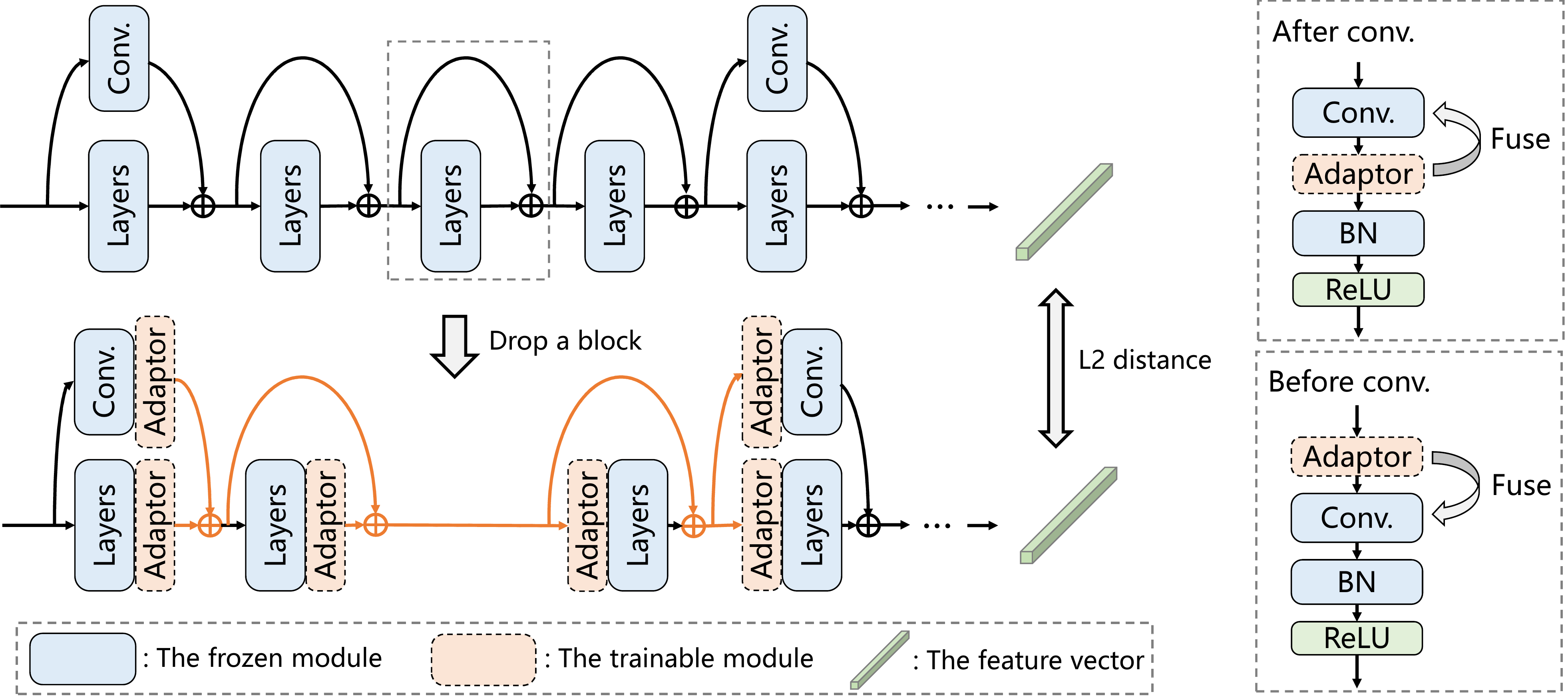}
	\caption{Illustration of our method for determining which block to drop. One block is dropped, and adaptors are inserted around the dropped position to recover from dropping this block. Then, the L2 distance of pruned/original networks' features are computed as the \emph{recoverability} of this dropped block. Finally, we drop several blocks that are easily recovered to obtain the pruned network. }
	\label{fig:practise}
\end{figure*}

\subsection{The recoverability of the pruned model}

To further improve dropping blocks, we study how to choose the blocks to drop. Existing criteria perform poorly in the few-shot scenario, as they either require a large training set~\cite{DBP,e_ResNet} or fail to predict the finetuned accuracy~\cite{CURL}. We argue that an effective metric should be consistent with the finetuned accuracy. But, as discussed in Sec.~\ref{sec:loss-convex}, it has a complex relationship to the convexity of loss landscape, initial accuracy, etc. In addition, traditional pruning steps in Fig.~\ref{fig:vanilla_FSP} will change the model's optimization landscape. Hence, it is inferior to simply evaluate the property of $f\leftrightarrow f'$ to indicate the finetuning of $g\leftrightarrow g'$. Instead, we will propose an algorithm, \Practise, to efficiently evaluate the difficulty of finetuning. In short, \Practise inserts adaptor layers around the pruning structure and optimizes the adaptor \emph{with other parts fixed}. Surprisingly, for few-shot finetuning, training only adaptors achieves almost the same optimization properties as training all parameters end-to-end.

First, we propose a new concept called \emph{recoverability}, which measures the ability of a pruned model to recover accuracy. As shown in Fig.~\ref{fig:metric}, we will take the `layer1.1' block as an example. Simply dropping this block results in a 54.3\% Top-1 error. Both KL-divergence (`CURL') and `L2 distance' variations before/after block removal are large, showing the model's outputs indeed change dramatically. But, the accuracy is recovered very effectively by finetuning with even a tiny training set. Finetuning is ignored in existing criteria, but it is finetuned accuracy that directly determines the compression quality. To overcome this difficulty, we propose a method to evaluate the recoverability of each dropped block. As illustrated in Fig.~\ref{fig:metric}, ours (`PRACTISE') enjoys a high consistency with the Top-1 error of the finetuned model.

Fig.~\ref{fig:practise} presents our method. Given the original model $\mathcal{M}_O$, the pruned model $\mathcal{M}_{P(\mathcal{B}_i)}$ is obtained by dropping a block $\mathcal{B}_i$. We insert adaptors in the positions connected to this block. These adaptors are conv. layers with kernel size $1\times 1$ and are placed before/after the original conv. layers according to different positions. For blocks in front of the dropped block, adaptors are inserted after conv. layers. On the contrary, adaptors are inserted before conv. layers. Because convolutions are linear, all adaptors can be fused in neighboring conv. layers while keeping the network functionality unchanged. That is, these adaptors will not become overhead to the pruned model. We use the tiny training set to optimize these adaptors to minimize the distance of original/pruned models' features (i.e., feature mimicking). The recoverability is calculated as
\begin{equation}
  \label{eq:recoverability}
  \mathcal{R}(\mathcal{B}_i) = \min_{\alpha} \mathbb{E}_{x\sim p(x)}\| \mathcal{M}_O(x;\theta) - \mathcal{M}_{P(\mathcal{B}_i)}(x;\theta\setminus b_i, \alpha)\|^2_F\,,
\end{equation}
where $\theta$ means parameters in the original model, and $\setminus b_i$ means excluding the parameters in the dropped block $\mathcal{B}_i$, and $\alpha$ denotes parameters in adaptors. This metric, as Fig.~\ref{fig:metric} shows, estimates the recoverability of each block very well.

\begin{figure*}
  \centering
 \subcaptionbox{\label{fig:FT_f} $\mathcal{M}_O \leftrightarrow \mathcal{M}_{P(B)}$}{\includegraphics[width=0.24\linewidth]{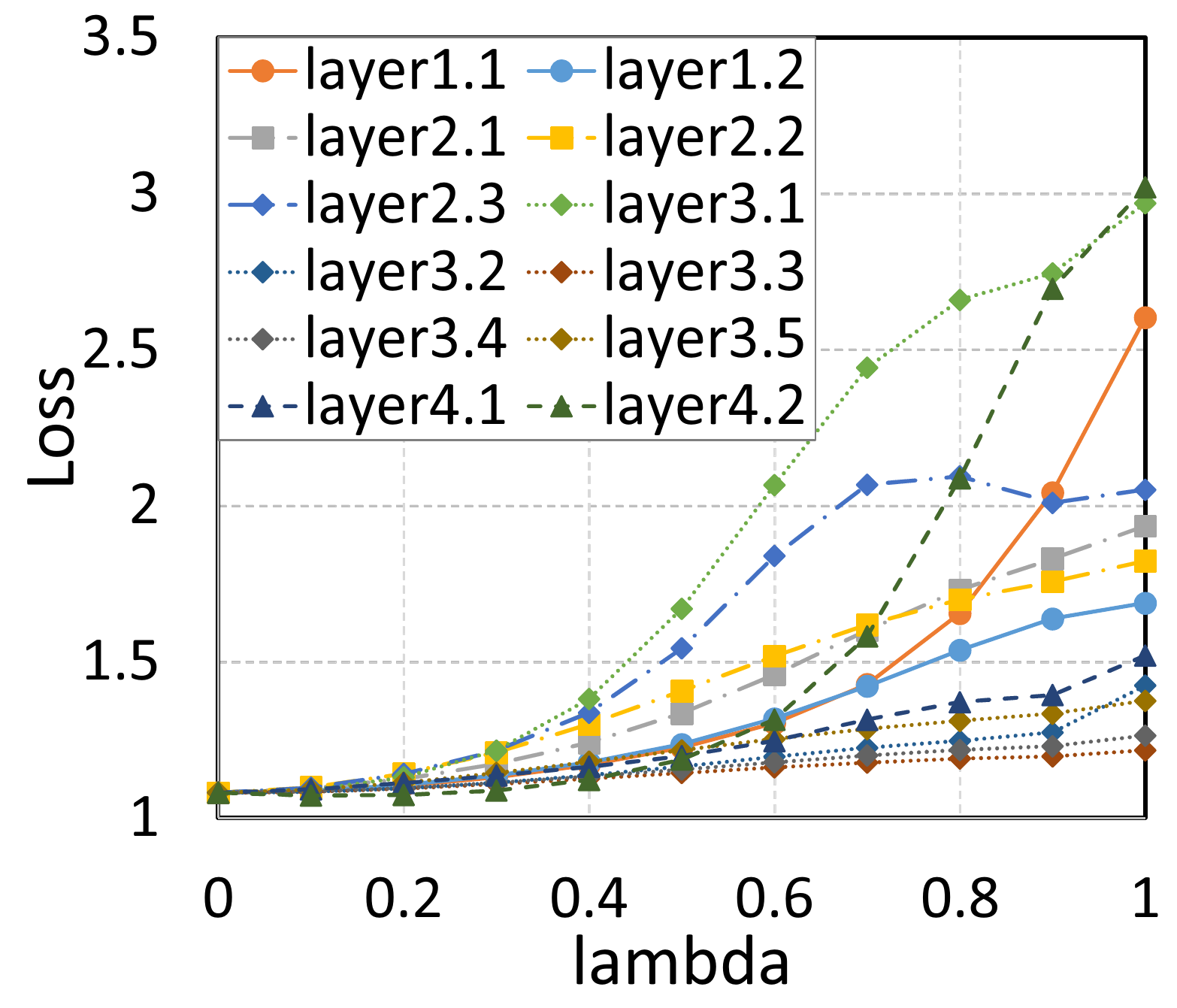}}
  \subcaptionbox{\label{fig:FT_g}$\mathcal{M}'_{P(B)} \leftrightarrow \mathcal{M}_{P(B)} $}{\includegraphics[width=0.24\linewidth]{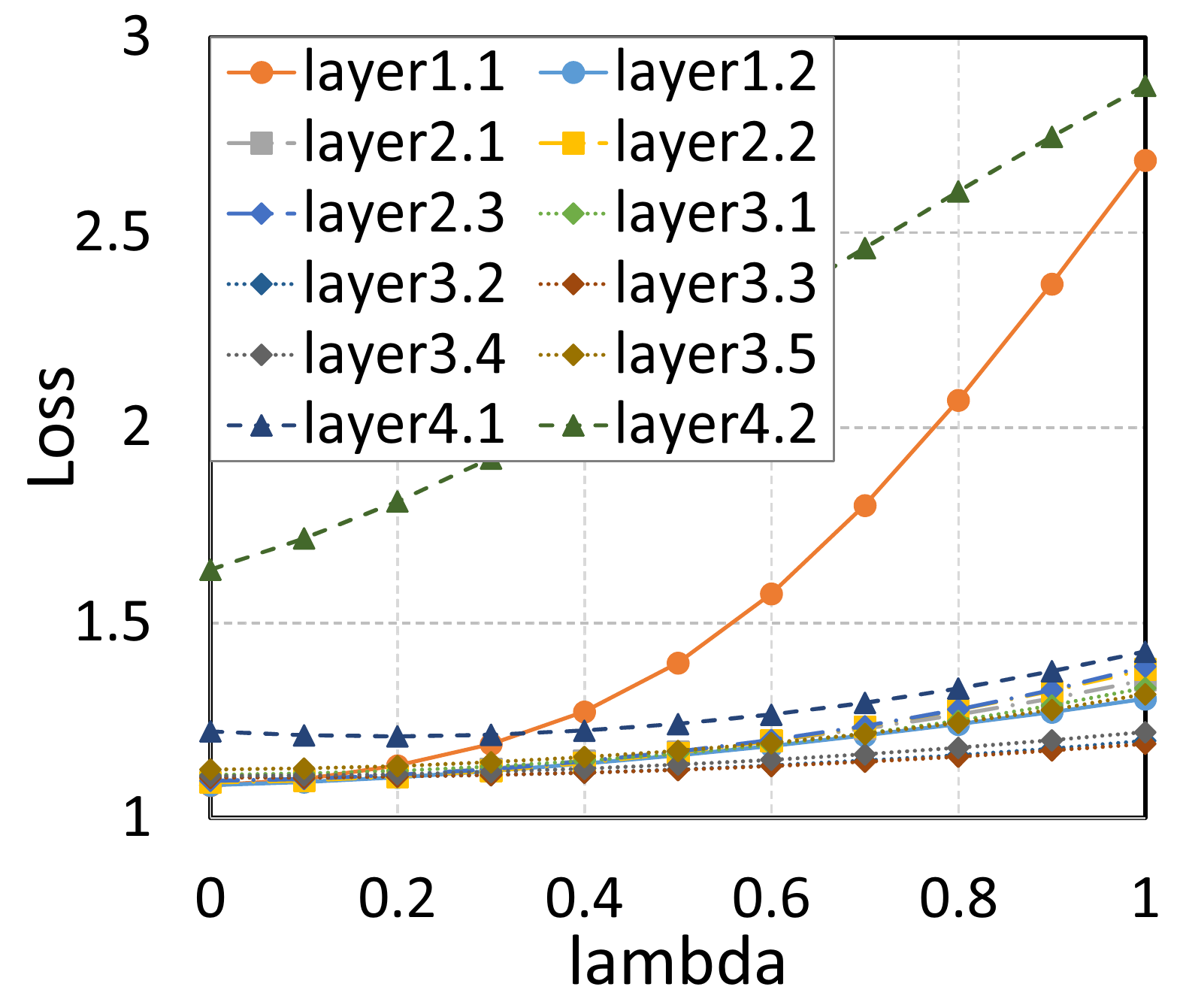}}
 \subcaptionbox{\label{fig:Practise_gh}$\mathcal{M}_{P(B)+A'} \leftrightarrow \mathcal{M}_{P(B)+A} $}{\includegraphics[width=0.24\linewidth]{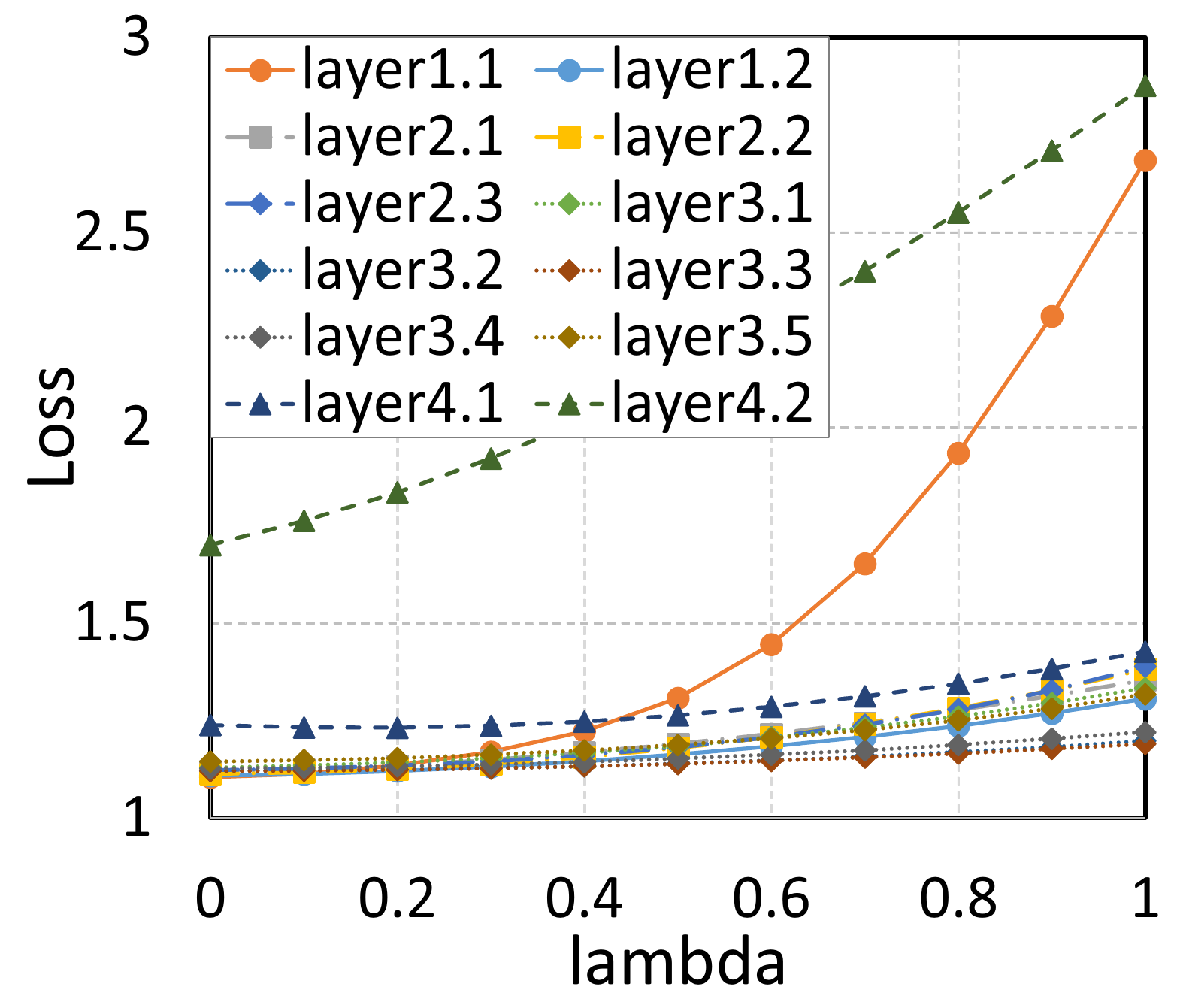}}
 \subcaptionbox{\label{fig:Practise_gg}$\mathcal{M}_{P(B)+A'} \leftrightarrow \mathcal{M}'_{P(B)} $}{\includegraphics[width=0.24\linewidth]{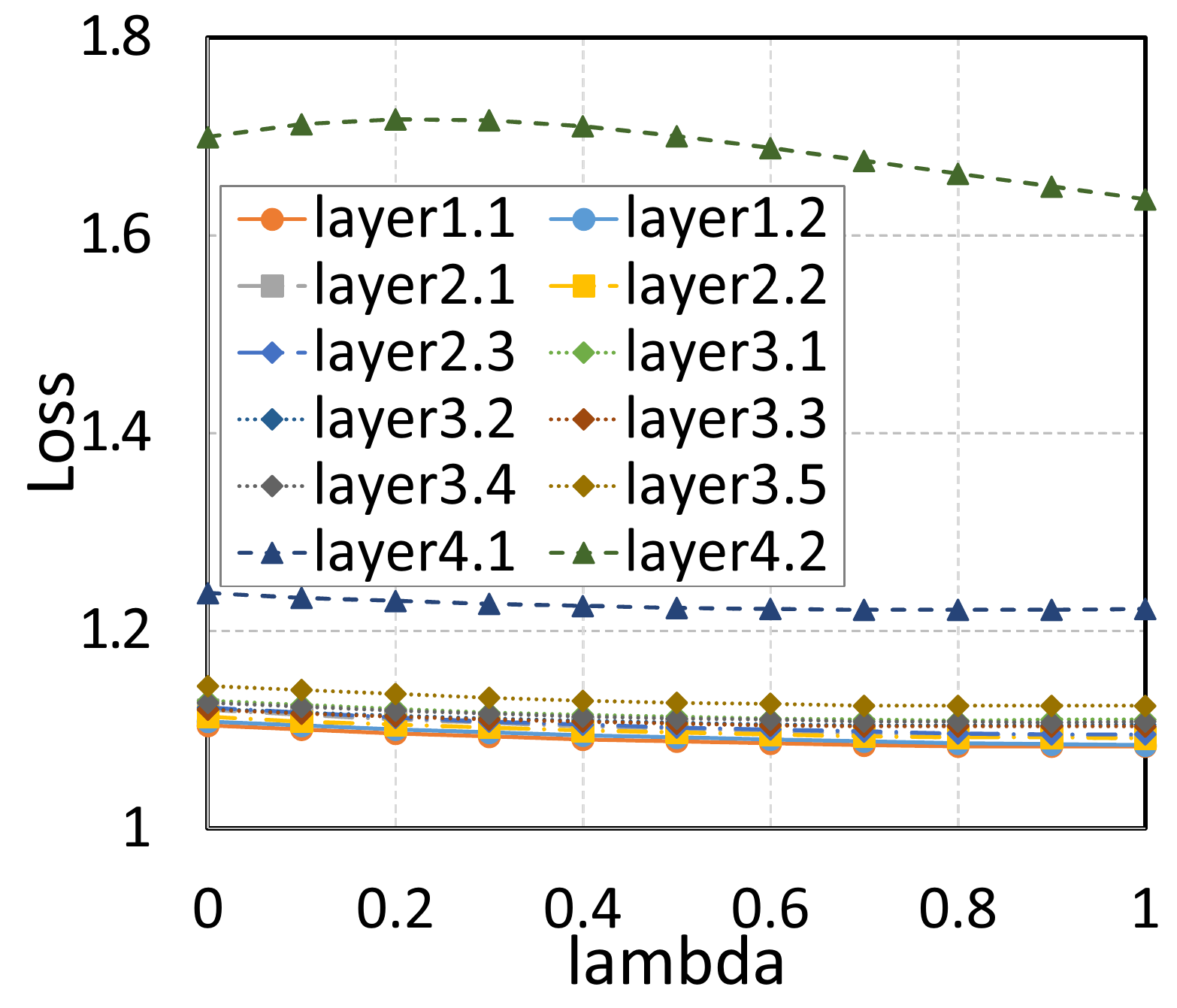}}
\caption{The loss of interpolated models between different pruned models. Each curve denotes pruning one block. Given the original model $\mathcal{M}_O$, the pruned model $\mathcal{M}_{P(B)}$ is obtained by dropping the block $B$, then finetuned on a tiny set to $\mathcal{M}'_{P(B)}$.  (\protect\subref{fig:FT_f}) and (\protect\subref{fig:FT_g}) compare these models' loss landscapes. Due to the change of architecture, the loss landscapes also change significantly, as (\protect\subref{fig:FT_f}) and (\protect\subref{fig:FT_g}) are significantly different. Note that \emph{$\mathcal{M}'_{P(B)}$ is desired target}. By inserting adaptors to get model $\mathcal{M}_{P(B)+A}$, our \Practise only trains these adaptors to achieve $\mathcal{M}_{P(B)+A'}$. \emph{optimizing adaptors almost perfectly achieves the desired target curve}---(\protect\subref{fig:FT_g}) and (\protect\subref{fig:Practise_gh}) aligns precisely. In particular, (\protect\subref{fig:Practise_gg}) reveals that the finetuned models from these two training methods lie in the same optimization basin without obvious loss barriers. }
\label{fig:FT_practise} 
\end{figure*}

\begin{figure}
	\centering
	\includegraphics[width=0.6\linewidth]{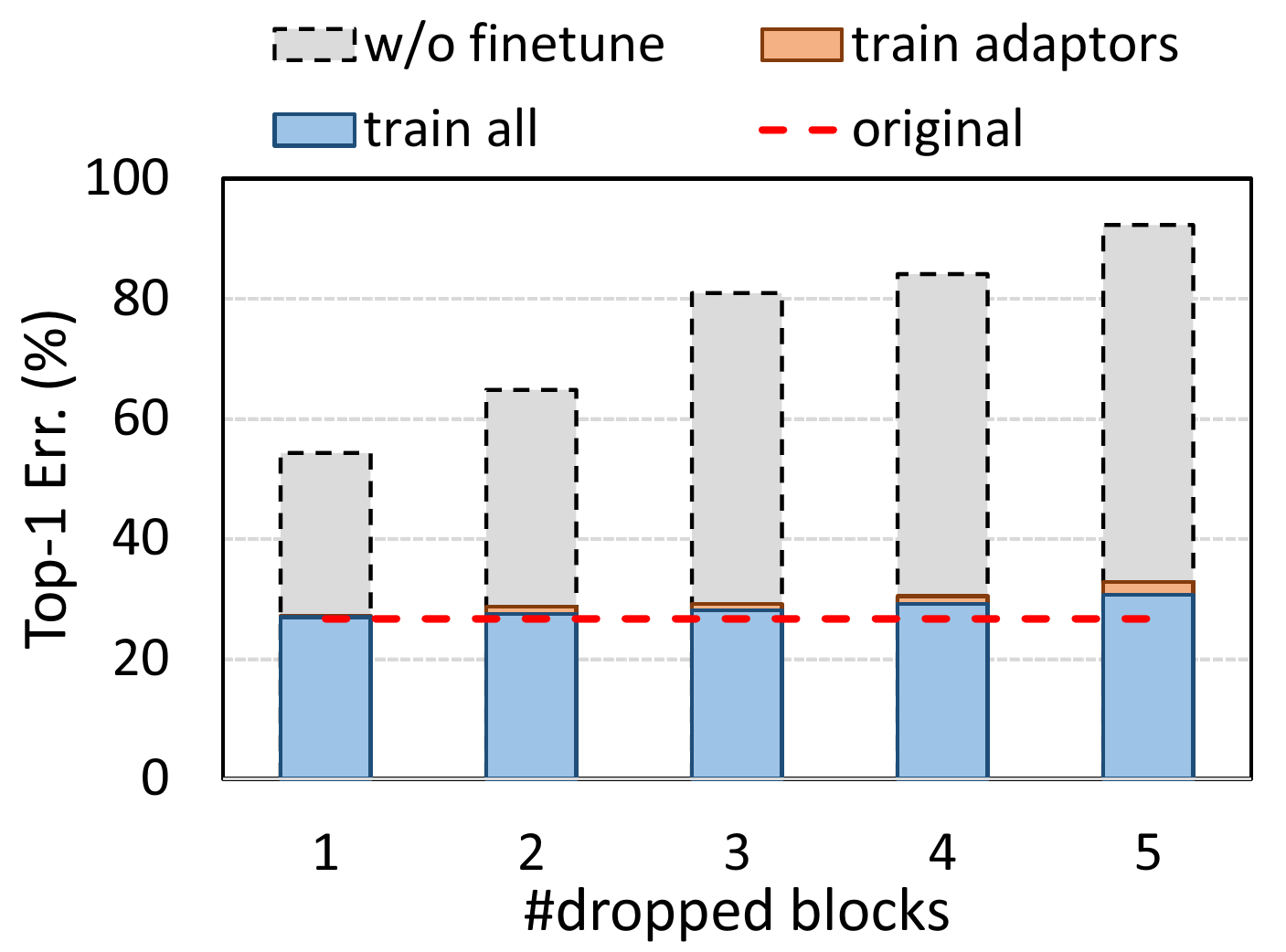}
	\caption{Comparison of training all and only adaptors. ResNet-34 was pruned by dropping different numbers of blocks with only 500 training samples. Top-1 error (\%) was evaluated on the ImageNet-1k validation set. Note that the Top-1 error of the original model is 26.98\%. }
	\label{fig:drop-5-blocks}
\end{figure}
	
Its success aligns well with our hypothesis and theory, as shown in Fig.~\ref{fig:FT_practise}. Comparing Fig.~\ref{fig:FT_f} and \ref{fig:FT_g}, we conclude that pruning architecture (the second step in Fig.~\ref{fig:vanilla_FSP}) changes the loss landscape. On the contrary, optimizing only adaptors ($\mathcal{M}_{P(B)+A'}\leftrightarrow \mathcal{M}_{P(B)+A}$) performs almost the same as finetuning the entire model ($\mathcal{M}'_{P(B)}\leftrightarrow \mathcal{M}_{P(B)}$). As demonstrated in Fig.~\ref{fig:Practise_gg}, the finetuned models $\mathcal{M}'_{P(B)}$ and $\mathcal{M}_{P(B)+A'}$ even lie in the same optimization basin without obvious loss barriers. In other words, \emph{optimizing only adaptors is almost the same as the end-to-end optimization}. They give strong empirical evidences that \Practise indeed finds recoverability precisely. In Fig.~\ref{fig:Practise_gh}, dropping ``layer1.1'' results in the most convex curve, which means it is easy to recover the accuracy by few samples. In contrast, ``layer4.2'' suffers from a flat curve. With a low initial accuracy, it struggles with poor performance. 

Fig.~\ref{fig:drop-5-blocks} further examines the outcome when we drop more blocks. It demonstrates that these adaptors recover most accuracy loss and achieve accuracy on-par with finetuning all parameters in the few-shot scenario. 

Our method is efficient, too. Thanks to the limited computations and parameters in adaptors, calculating the recoverability by Eq.~\ref{eq:recoverability} requires only a few training samples and very short training time.

The other factor for dropping blocks we must take care of is the latencies of different blocks. With the same recoverability, the block with a higher latency should have a higher priority to drop. To this end, we calculate the acceleration ratio of the block $\mathcal{B}_i$ by 
\begin{equation}
  \label{eq:latency-ratio}
  \tau(\mathcal{B}_i) = \frac{lat_{\mathcal{M}_O} - lat_{\mathcal{M}_{P(\mathcal{B}_i)}}}{lat_{\mathcal{M}_O}}\,,
\end{equation}
where $lat$ denotes the latency. Finally, we define the pruning score for each block $\mathcal{B}_i$ as 
\begin{equation}
  \label{eq:score}
  s(\mathcal{B}_i) = \frac{\mathcal{R}(\mathcal{B}_i)}{\tau(\mathcal{B}_i)}\,.
\end{equation}
Our pruning score considers both recoverability and latency of each block. A lower score means the block has a higher priority to drop. We compute the pruning score for each block and drop the top $k$ blocks with smallest scores.

\subsection{Recover the accuracy of the pruned model}
\label{sec:FM}

Once the pruned model's structure is determined, the last problem is how to recover the accuracy. The advantages of MiR has been proved in Sec.~\ref{sec:theory}. Hence, following MiR, we finetune the pruned model by minimizing
\begin{equation}
  \label{eq:ft-loss}
  \mathcal{L} = \| \mathcal{M}_O(x;\theta_O) - \mathcal{M}_P(x;\theta_P)\|^2_F\,,
\end{equation}
where $\theta_O$ denotes the frozen parameters of the original model and $\theta_P$ denotes the trainable parameters of the pruned model. 

\begin{algorithm}[t]
  \caption{\Practise}
  \label{alg:practise}
  \KwIn{The original model $\mathcal{M}_O$, the number of dropped blocks $k$, the tiny training data $\mathcal{D}_{\mathcal{T}}$}
  Test the latency of $\mathcal{M}_O$\;
  \For{each block $\mathcal{B}_i$}
  {
    Drop $\mathcal{B}_i$ to obtain the pruned model $\mathcal{M}_{P(\mathcal{B}_i)}$\;
    Test latency of $\mathcal{M}_{P(\mathcal{B}_i)}$ and find $\tau(\mathcal{B}_i)$ (Eq.~\ref{eq:latency-ratio})\;
    Insert adaptors\;
    Compute $\mathcal{R}(\mathcal{B}_i)$ with $\mathcal{D}_{\mathcal{T}}$ (Eq.~\ref{eq:recoverability})\;
    Compute the score $s(\mathcal{B}_i)$ (Eq.~\ref{eq:score})\;
    Add $\mathcal{B}_i$ back and remove all adaptors;
  }
  Choose the top $k$ blocks with the smallest scores\;
  Drop these $k$ blocks to obtain $\mathcal{M}_P$\;
  Finetune $\mathcal{M}_P$ with $\mathcal{D}_{\mathcal{T}}$ by minimizing $\mathcal{L}$ (Eq.~\ref{eq:ft-loss})\;
  \textbf{return} The pruned model $\mathcal{M}_P$
\end{algorithm}

Overall, the whole algorithm of \Practise is presented in Alg.~\ref{alg:practise}. Our \Practise enjoys \emph{zero} extra hyperparameters. With feature mimicking, we accelerate models \emph{without} using training labels.

\Practise can even work in data-free scenarios. One choice is to adopt the synthesized images from DFKD~\cite{DI} as training images. Most existing DFKD work~\cite{DI,MixMix} adopt filter pruning, whereas \Practise improves the latency-accuracy performance by a significant margin. That helps a lot in data-free scenarios. Another choice is to collect out-of-domain data. On the other hand, with a large amount of out-of-domain images, the accuracy of the pruned network is even close to that of using original training images. That demonstrates the high generalization ability of \PractiseNoSpace. 

\section{Experimental Results}
\label{sec:exp}

\begin{table*}
  \caption{Top-1/Top-5 validation accuracy (\%) on ImageNet-1k for pruning ResNet-34. All models reduced $16\%$ latency with 50, 100, 500, and 1000 training samples. Previous methods pruned filters within the residual block. Top-1/Top-5 accuracy of the original ResNet-34 are $73.31\%/91.42\%$.}
  \label{tab:ImageNet-ResNet34-highlan-4data}
	\centering
  \small
	\begin{tabular}{c|c|cccc}
		\toprule
		Method & Latency (ms) & 50	& 100 & 500	& 1000 \\
		\midrule
    BP (filter) & 35.1 (15.8\%$\downarrow$) & $39.0_{\pm 1.41}/68.9_{\pm 1.17}$ & $41.0_{\pm 0.33}/70.5_{\pm 0.66}$ & $51.8_{\pm 0.30}/78.1_{\pm 0.38}$ & $57.8_{\pm 0.30}/81.5_{\pm 0.18}$\\

    BP (block) & \textbf{34.9} (\textbf{16.3}\%$\downarrow$) & $\textbf{66.5}_{\pm 0.81}/\textbf{78.4}_{\pm 0.44}$ & $\textbf{66.8}_{\pm 0.23}/\textbf{87.7}_{\pm 0.23}$ & $\textbf{68.6}_{\pm 0.18}/\textbf{88.8}_{\pm 0.09}$ & $\textbf{69.8}_{\pm 0.12}/\textbf{89.3}_{\pm 0.07}$\\
    \midrule

    KD~\cite{KD} & 35.1 (15.8\%$\downarrow$) & $44.5_{\pm 1.20}/72.3_{\pm 0.87}$ & $46.4_{\pm 0.34}/74.0_{\pm 0.58}$ & $54.7_{\pm 0.26}/79.7_{\pm 0.19}$ & $57.9_{\pm 0.21}/81.6_{\pm 0.12}$\\

    FSKD~\cite{FSKD} & 35.1 (15.8\%$\downarrow$) & $45.3_{\pm 0.77}/71.5_{\pm 0.62}$ & $51.2_{\pm 0.30}/76.8_{\pm 0.23}$ & $57.6_{\pm 0.21}/81.6_{\pm 0.15}$ & $59.4_{\pm 0.13}/82.7_{\pm 0.06}$\\

    CD~\cite{CD} & 35.1 (15.8\%$\downarrow$) & $56.2_{\pm 0.37}/80.8_{\pm 0.31}$ & $59.1_{\pm 0.22}/82.8_{\pm 0.11}$ & $63.7_{\pm 0.18}/86.0_{\pm 0.05}$ & $64.4_{\pm 0.03}/86.3_{\pm 0.07}$\\

    MiR~\cite{MiR} & 35.1 (15.8\%$\downarrow$) & $64.1_{\pm 0.10}/86.3_{\pm 0.11}$ & $65.1_{\pm 0.19}/87.0_{\pm 0.11}$ & $67.0_{\pm 0.09}/88.1_{\pm 0.07}$ & $67.8_{\pm 0.06}/88.5_{\pm 0.02}$\\

    \Practise & \textbf{34.9} (\textbf{16.3}\%$\downarrow$) & $\textbf{70.3}_{\pm 0.16}/\textbf{89.6}_{\pm 0.06}$ & $\textbf{71.5}_{\pm 0.74}/\textbf{90.3}_{\pm 0.37}$ & $\textbf{72.5}_{\pm 0.04}/\textbf{90.9}_{\pm 0.03}$ & $\textbf{72.5}_{\pm 0.05}/\textbf{91.0}_{\pm 0.02}$ \\
		\bottomrule
	\end{tabular}
\end{table*}

\begin{table*}
  \caption{Top-1/Top-5 validation accuracy (\%) on ImageNet-1k for pruning ResNet-34. Our model reduced $22.1\%$ latency. Previous methods pruned filters both inside and outside the residual connection (i.e., `residual' in Fig.~\ref{fig:latency-flops}). Top-1/Top-5 accuracy of the original ResNet-34 are $73.31\%/91.42\%$.}
  \label{tab:ImageNet-ResNet34-lowlan-4data}
	\centering
  \small
	\begin{tabular}{c|c|cccc}
		\toprule
		Method & Latency (ms) & 50	& 100 & 500	& 1000 \\
		\midrule
    BP (filter) & $33.8\,(18.9\%\downarrow)$ & $24.2_{\pm 0.92}/52.7_{\pm 1.36}$ & $27.6_{\pm 0.41}/56.7_{\pm 0.62}$ & $42.9_{\pm 0.28}/70.5_{\pm 0.27}$ & $51.2_{\pm 0.32}/76.5_{\pm 0.16}$\\

    BP (block) & $\textbf{32.5}\,(\textbf{22.1}\%\downarrow)$ & $\textbf{60.6}_{\pm 0.62}/\textbf{83.5}_{\pm 0.42}$ & $\textbf{61.6}_{\pm 0.31}/\textbf{84.3}_{\pm 0.36}$ & $\textbf{65.0}_{\pm 0.19}/\textbf{86.5}_{\pm 0.20}$ & $\textbf{66.8}_{\pm 0.18}/\textbf{87.5}_{\pm 0.13}$\\
    \midrule

    KD~\cite{KD} & $33.8\,(18.9\%\downarrow)$ & $30.1_{\pm 0.69}/57.7_{\pm 1.10}$ & $33.1_{\pm 0.43}/61.0_{\pm 0.53}$ & $45.7_{\pm 0.26}/72.2_{\pm 0.25}$ & $50.5_{\pm 0.29}/75.9_{\pm 0.23}$\\

    FSKD~\cite{FSKD} & $33.8\,(18.9\%\downarrow)$ & $31.1_{\pm 0.90}/56.5_{\pm 1.10}$ & $36.6_{\pm 0.44}/63.1_{\pm 0.46}$ & $42.8_{\pm 0.49}/69.1_{\pm 0.58}$ & $44.9_{\pm 0.20}/70.5_{\pm 0.29}$\\

    MiR~\cite{MiR} & $33.8\,(18.9\%\downarrow)$ & $59.9_{\pm 0.30}/83.2_{\pm 0.31}$ & $62.1_{\pm 0.22}/84.8_{\pm 0.18}$ & $65.4_{\pm 0.07}/87.0_{\pm 0.03}$ & $66.6_{\pm 0.05}/87.7_{\pm 0.04}$\\

    \Practise & $\textbf{32.5}\,(\textbf{22.1}\%\downarrow)$ & $\textbf{68.0}_{\pm 1.36}/\textbf{88.2}_{\pm 0.77}$ & $\textbf{70.4}_{\pm 0.42}/\textbf{89.7}_{\pm 0.23}$ & $\textbf{71.8}_{\pm 0.07}/\textbf{90.5}_{\pm 0.02}$ & $\textbf{71.9}_{\pm 0.05}/\textbf{90.6}_{\pm 0.04}$ \\
		\bottomrule
	\end{tabular}
\end{table*}

\begin{table*}
  \caption{Top-1 validation accuracy (\%) on ImageNet-1k for pruning ResNet-34. The model was pruned at three different compression ratios with 500 training samples. Previous methods prune filters by CD-style, while we drop blocks. The Top-1 accuracy and the latency of the original ResNet-34 are $73.31\%$ and $41.7$ ms, respectively. }
  \label{tab:ImageNet-ResNet34-Top1-LA-500}
	\centering
  \small
	\begin{tabular}{c|r@{.}lr@{.}lr@{.}lr@{.}lr@{.}lr@{.}l}
		\toprule
		Latency (ms) & \multicolumn{2}{c}{BP (filter)} & \multicolumn{2}{c}{KD~\cite{KD}} & \multicolumn{2}{c}{FSKD~\cite{FSKD}}	& \multicolumn{2}{c}{CD~\cite{CD}} & \multicolumn{2}{c}{MiR~\cite{MiR}} & \multicolumn{2}{c}{\Practise}	\\
		\midrule
    32.5 (22.1\%$\downarrow$) & $47$&$54_{\pm 0.41}$ & $49$&$34_{\pm 0.25}$ & $33$&$19_{\pm 0.60}$ & $59$&$65_{\pm 0.12}$ & $68$&$14_{\pm 0.04}$ & $\mathbf{71}$&$\mathbf{75}_{\pm 0.07}$ \\

    34.9 (16.3\%$\downarrow$) & $58$&$94_{\pm 0.36}$ & $61$&$01_{\pm 0.24}$ & $62$&$56_{\pm 0.13}$ & $68$&$17_{\pm 0.07}$ & $70$&$53_{\pm 0.10}$ & $\mathbf{72}$&$\mathbf{52}_{\pm 0.04}$ \\

    38.3 \,\,\,(8.2\%$\downarrow$) & $65$&$02_{\pm 0.30}$ & $67$&$22_{\pm 0.18}$ & $69$&$59_{\pm 0.09}$ & $71$&$12_{\pm 0.06}$ & $71$&$95_{\pm 0.07}$ & $\mathbf{73}$&$\mathbf{04}_{\pm 0.06}$ \\

		\bottomrule
	\end{tabular}
\end{table*}

In this section, we present \Practise results on both few-shot and data-free compression. Following previous work~\cite{FSKD,CD,MiR}, ResNet-34~\cite{ResNet} and MobileNetV2~\cite{mobilenet} will be pruned on tiny training sets of ImageNet-1k~\cite{imagenet}. Then, to test the generalization ability of \PractiseNoSpace, we will prune ResNet-50 with synthesized images and out-of-domain data, respectively. Finally, ablation studies are conducted .

\textbf{Implementation details.} As presented in Alg.~\ref{alg:practise}, \Practise requires computing the latency and recoverability of each pruned model $\mathcal{M}_{P(\mathcal{B}_i)}$, then finetuning the pruned model $\mathcal{M}_{P}$. In this paper, we only consider dropping the block with the same input and output dimensionality. For the latency, we tested the model with $64\times 3\times 224\times 224$ input 500 times and record the mean latency. All latency numbers were tested on the same computer with an NVIDIA TITAN Xp GPU. To compute recoverability, we used SGD with batch size $64$ to optimize adaptors by 1000 iterations. The initial learning rate was $0.02$ and decreased by a factor of 10 per 40\% iterations. When the size of training data $\mathcal{D}_{\mathcal{T}}$ is less than $64$, batch size was the data size. For the final finetuning, all parameters in the pruned network were updated by SGD to minimize Eq.~\ref{eq:ft-loss} (i.e., feature mimicking) with 2000 iterations by default. Batch size and learning rate schedules were the same as those for adaptors. We used the data argumentation strategy supplied by PyTorch official examples, which is the same as that of previous work~\cite{FSKD,CD,MiR}. All experiments were conducted with PyTorch~\cite{pytorch}.

To compare with other few-shot pruning methods, we are mainly concerned about the latency-accuracy tradeoff of the pruned model. We tested the latency of networks pruned by previous methods and directly cite their accuracy numbers. The results of \Practise were run by five times with different sampled tiny training sets. We report the mean accuracy along with standard deviation. 

\subsection{Different amounts of training data}

We compare \Practise with state-of-the-art few-shot pruning methods. Following previous work, we prune ResNet-34 with different amounts of training images on ImageNet-1k. Table~\ref{tab:ImageNet-ResNet34-highlan-4data} summarizes results. All previous methods pruned filters within residual blocks (cf. `normal' in Fig.~\ref{fig:latency-flops}), and both our \Practise and dropping block scheme achieve significantly better results. First, we removed the first few blocks to reach the latency goal and then simply finetuned the network with the cross-entropy loss. This simple baseline for dropping blocks, `BP (block)', has already outperformed previous methods. Note that this is our contribution, too, which reveals the advantage of dropping blocks in the few-shot scenario. \Practise further improves results. With similar latency reduction, it dramatically outperforms previous SOTA by an average of $5.7\%$ Top-1 accuracy on ImageNet-1k. Table~\ref{tab:ImageNet-ResNet34-lowlan-4data} compares these methods with a larger latency reduction. \Practise surpasses MiR by a significant margin again, on average $7.0\%$ Top-1 accuracy. Note that our model is faster than previous ones by $1.3$ ms, too.

\subsection{Different acceleration ratios}

Table~\ref{tab:ImageNet-ResNet34-Top1-LA-500} compares \Practise with previous methods for different acceleration ratios. For $8.2\%$ latency reduction, \Practise outperforms MiR by $1.1\%$ Top-1 accuracy. In further reducing latency by $22.1\%$, \Practise surpasses MiR by $3.6\%$ Top-1 accuracy. That indicates \Practise enjoys higher accuracy than others when the acceleration ratio becomes larger. Fig.~\ref{fig:top1-latency} presents curves for the latency-accuracy tradeoffs of different methods. Our \Practise outperforms previous methods dominantly. 

\subsection{The data-latency-accuracy tradeoff}

\begin{figure}
	\centering
	\includegraphics[width=0.7\linewidth]{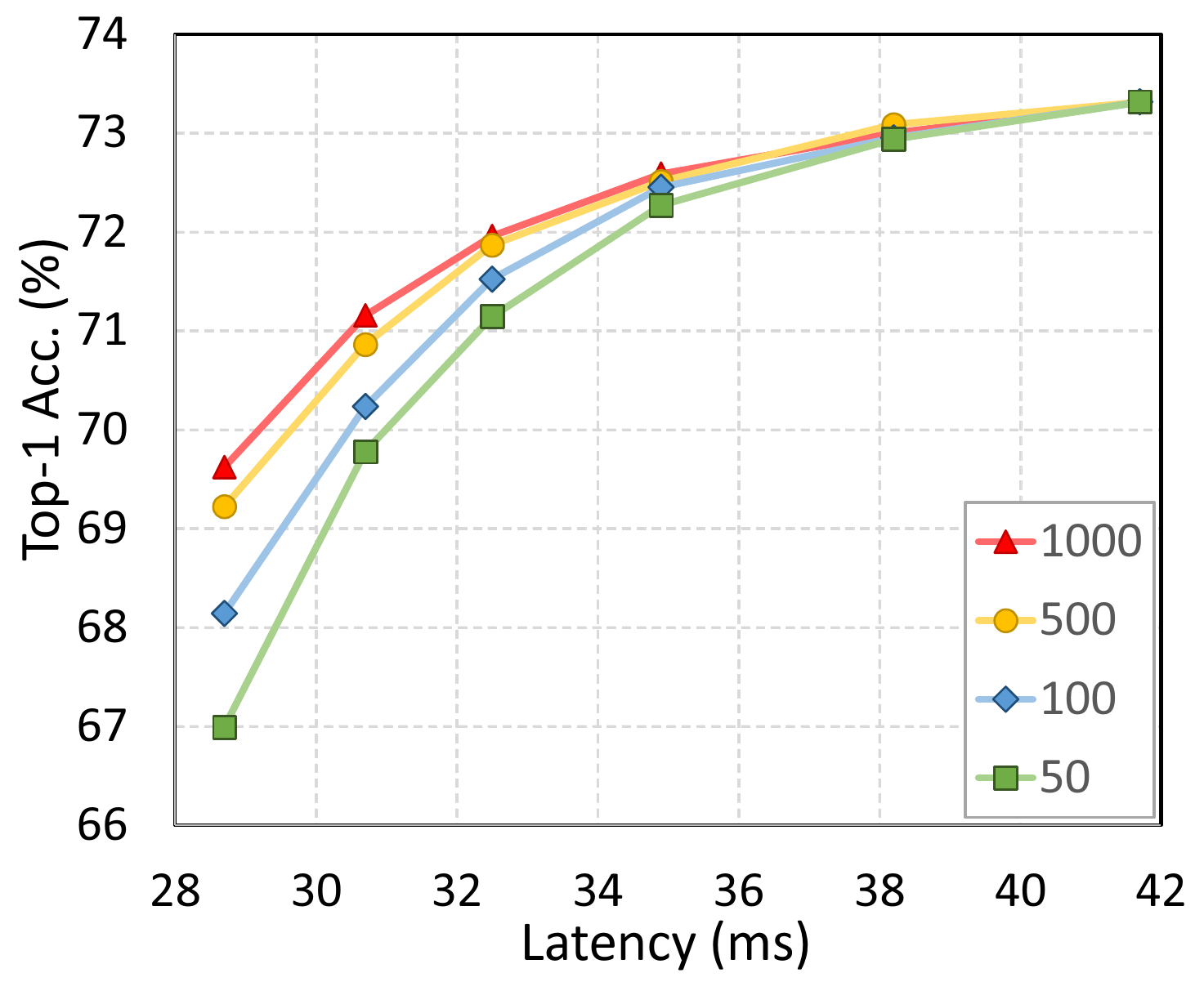}
	\caption{The data-latency-accuracy tradeoff. ResNet-34 was pruned by \Practise on ImageNet-1k. Numbers in the legend mean different amounts of training images. }
	\label{fig:data_latency_acc}
\end{figure}

Next, for a better understanding of the data-latency-accuracy tradeoff in the few-shot compression scenario, we pruned ResNet-34 by \Practise with different latency reductions and various numbers of training images. Fig.~\ref{fig:data_latency_acc} presents the results. With less latency reduction, the accuracies for different amounts of training data are comparable. But for a large latency reduction, it is challenging to recover the accuracy by only a tiny training set. And the accuracy gap becomes large w.r.t. different amounts of training data.  

\subsection{Results on MobileNetV2}

\begin{table}
  \caption{Comparison of \Practise and few-shot pruning methods on ImageNet-1k. MobileNetV2 was pruned with 500 training samples. }
  \label{tab:ImageNet-MobileNetV2}
	\centering
  \small
	\begin{tabular}{c|cc}
		\toprule
		Method & Latency (ms) & Top-1/Top-5	\\
		\midrule
    Original & \textcolor{lightgray}{37.6} & \textcolor{lightgray}{71.9/90.3} \\
    \midrule
    BP (filter) & $31.5\,(16.2\%\downarrow)$ & $45.0_{\pm 0.34}/71.8_{\pm 0.38}$ \\

    KD~\cite{KD} & $31.5\,(16.2\%\downarrow)$ & $48.4_{\pm 0.34}/73.9_{\pm 0.32}$ \\

    MiR~\cite{MiR} & $31.5\,(16.2\%\downarrow)$ & $67.6_{\pm 0.05}/87.9_{\pm 0.04}$ \\

    \Practise & $\textbf{30.4}\,(\textbf{19.1}\%\downarrow)$ & $\textbf{69.3}_{\pm 0.05}/\textbf{88.9}_{\pm 0.05}$ \\
    \midrule
    BP (filter) & $34.1\,\,\,(9.3\%\downarrow)$ & $55.5_{\pm 0.16}/80.3_{\pm 0.26}$ \\

    KD~\cite{KD} & $34.1\,\,\,(9.3\%\downarrow)$ & $59.1_{\pm 0.17}/82.5_{\pm 0.15}$ \\

    MiR~\cite{MiR} & $34.1\,\,\,(9.3\%\downarrow)$ & $69.7_{\pm 0.04}/89.2_{\pm 0.03}$ \\

    \Practise & $\textbf{31.9}\,(\textbf{15.2}\%\downarrow)$ & $\textbf{70.3}_{\pm 0.03}/\textbf{89.5}_{\pm 0.03}$ \\
		\bottomrule
	\end{tabular}
\end{table}

MobileNetV2~\cite{mobilenet} is a lightweight model and is popularly applied on mobile devices. It has 10 blocks that can be dropped by \PractiseNoSpace. Table~\ref{tab:ImageNet-MobileNetV2} summarizes MobileNetV2 results. \Practise outperforms previous methods by a significant margin. Compared with MiR, \Practise obtains pruned models with both lower latency and higher accuracy. 

\subsection{Train with synthesized/out-of-domain images}

\begin{table}[t]
  \caption{Comparison of \Practise and data-free methods on ImageNet-1k. Latency (ms) and Top-1 validation accuracy (\%) on ImageNet-1k are reported. * denotes the method reimplemented by ourselves. }
  \label{tab:ADI}
	\centering
  \setlength{\tabcolsep}{2pt}
  \small
	\begin{tabular}{c|cccc}
		\toprule
		Network & Method & Pruning & Latency & Top-1	\\
		\midrule
    \multirow{6}*{ResNet-50} & 
    Original & & \textcolor{lightgray}{83.8} & \textcolor{lightgray}{76.1} \\
    % \cmidrule(r){2-5}
    & DI~\cite{DI} & filter & - & $72.0$ \\
    & MixMix~\cite{MixMix} & filter & - & $69.8$ \\
    & ADI~\cite{DI} & filter & - & $73.3$ \\
    & ADI*~\cite{DI} & filter & 79.9 \,\,\,(4.7\%$\downarrow$) & $73.5$ \\
    % & \Practise & block & 62.7 (25.2\%$\downarrow$) & $73.7$ \\
    & \Practise & block & \textbf{66.2} (\textbf{21.0}\%$\downarrow$) & \textbf{74.8} \\
    % & \Practise & block & 77.4 (15.5\%$\downarrow$) & $75.8$ \\
    \midrule
    \multirow{5}*{MobileNetV2} & 
    Original & & \textcolor{lightgray}{37.6} & \textcolor{lightgray}{71.9} \\ 
    % \cmidrule(r){2-5}
    & DI~\cite{DI} & filter & - & 15.3 \\
    & MixMix~\cite{MixMix} & filter & - & 42.5 \\
    & ADI*~\cite{DI} & filter & 30.8 (18.1\%$\downarrow$) & 62.8 \\
    & \Practise & block & \textbf{30.4} (\textbf{19.1}\%$\downarrow$) & \textbf{68.0} \\
		\bottomrule
	\end{tabular}
\end{table}

\begin{table}
  \caption{Top-1 validation accuracy (\%) on ImageNet-1k for different out-of-domain training datasets. ResNet-50 was reduced $21\%$ latency by \PractiseNoSpace. }
  \label{tab:Otherdata-ResNet50}
	\centering
  \small
	\begin{tabular}{c|ccccc}
		\toprule
		Dateset & 50 & 500 & 1000 & 5000 & All \\
		\midrule
    ImageNet~\cite{imagenet} & \textcolor{lightgray}{74.22} & \textcolor{lightgray}{74.58} & \textcolor{lightgray}{74.58} & \textcolor{lightgray}{75.14} & \textcolor{lightgray}{75.24} \\
    ADI~\cite{DI}  & 69.85 & 72.68 & 73.01 & 74.40 & 74.79 \\
    CUB~\cite{CUB} & 72.49 & 73.71 & 73.94 & 74.86 & 74.92 \\
    Place365~\cite{places365} & \textbf{72.80} & \textbf{74.10} & \textbf{74.18} & \textbf{75.05} & \textbf{75.21} \\
		\bottomrule
	\end{tabular}
\end{table}

In some extreme scenarios, not even a single original training sample is to be provided. Zero-shot pruning is required. Because \Practise does not need ground-truth labels, it is able to accelerate networks with images synthesized by data-free knowledge distillation methods. Table~\ref{tab:ADI} summarizes results. Note that most zero-shot pruning methods adopt filter pruning, which are inferior to pruning blocks as we have shown. We adopt synthesized images produced by ADI~\cite{DI} as the training set. For both ResNet-50 and MobileNetV2, our \Practise achieves higher Top-1 accuracy with more latency reductions. We advocate that we need to adopt dropping blocks for DFKD in the future.

Another choice for zero-shot pruning is to collect out-of-domain training images. \Practise is also robust to work with these data. Table~\ref{tab:Otherdata-ResNet50} presents results. The original ResNet-50 was trained on ImageNet-1k, and we pruned it on other datasets by \PractiseNoSpace. ADI consists of images synthesized by DeepInversion~\cite{DI}. CUB~\cite{CUB} contains images of birds with 200 categories. Place365~\cite{places365} consists of scene pictures. Our \Practise enjoys a high generalization ability to work with all these datasets. Another benefit of out-of-domain data is the unlimited number of images. We notice the accuracy is boosted by using more training samples and even close to that of using the entire original ImageNet-1k training set.

\subsection{Different criteria for dropping blocks}

\begin{table}[t]
  \caption{Top-1 validation accuracy (\%) on ImageNet-1k for pruning ResNet-34 with 500 images. We compare different criteria for dropping different numbers of blocks. }
  \label{tab:ImageNet-ResNet34-choose-block}
	\centering
  \small
	\begin{tabular}{c|ccccc}
		\toprule
		\#Dropped blocks & 1 & 2 & 3 & 4	& 5 \\
		\midrule
    CURL~\cite{CURL}  & 72.33 & 71.08 & 69.14 & 65.48 & 64.97 \\
    $\epsilon$-ResNet~\cite{e_ResNet} & 72.51 & 71.20 & 69.00 & 67.90 & 64.75 \\
    L2 distance       & 72.51 & 71.20 & 69.00 & 68.61 & 64.93 \\
    \Practise         & \textbf{73.02} & \textbf{72.52} & \textbf{71.92} & \textbf{70.86} & \textbf{69.35} \\
		\bottomrule
	\end{tabular}
\end{table}

Finally, we compare \Practise with other criteria for dropping blocks. Fig.~\ref{fig:metric} already shows results for removing only one block. Obviously, \Practise is better than others, and is very consistent with the finetuned models' accuracies. Most existing methods mainly measure the gap between the original network and the pruned network without finetuning. They neglect the recoverability of each dropped block, hence resulting in an inferior strategy for dropping blocks. Our \Practise pays attention to the recoverability and the acceleration ratio of each block. Therefore the pruned network enjoys higher accuracy and lower latency. 

Table~\ref{tab:ImageNet-ResNet34-choose-block} compares different criteria for dropping more blocks. As the number of dropped blocks increases, our \Practise outperforms others by even larger margins. To sum up, \Practise finds the best set of blocks to drop compared with other methods.

\section{Conclusions and Future Work}

This paper aims at accelerating networks with tiny training sets. We revealed a crucial observation in the few-shot compression: finetuning a pruned network is almost a convex problem. This conjecture successfully explains why related algorithms do not suffer from overfitting problems. We believe this finding makes significant progress in few-shot model compression. We further established a theory for feature mimicking. Compared to the vanilla method, it achieves lower parameter variances. Based on our theoretical results, for the first time, we revealed that dropping blocks is fundamentally better than previous filter-level pruning in this scenario, in terms of higher convexity and acceleration ratio. To determine which blocks to drop, we proposed a new concept, \emph{recoverability}, to measure the difficulty of recovering the pruned network's accuracy with few samples. Combining recoverability and feature mimicking, \Practise was proposed. It enjoys high latency-accuracy performance and is robust to deal with synthesized/out-of-domain images. Extensive experiments demonstrated that \Practise outperforms previous methods by a significant margin (on average 7\% Top-1 accuracy on ImageNet-1k for 22\% latency reduction).

Our conjecture has limitations such as it is confined to finetuning a pruned model. It needs future explorations for other scenarios like transfer learning, domain adaptation, etc. Our theory is proposed mainly for the discriminative model. How to distill the generative model is still an open problem. For network acceleration, how to compute the recoverability of other compression schemes is also an interesting problem. Recently, finetuning and accelerating a large pretrained network on downstream tasks are emerging as critical needs. Applying \Practise for tuning a pretrained model on downstream tasks is worth exploring, too.

\ifCLASSOPTIONcaptionsoff
  \newpage
\fi

\bibliographystyle{IEEEtran}
\bibliography{egbib}

\begin{IEEEbiography}[{\includegraphics[width=1in,height=1.25in,clip,keepaspectratio]{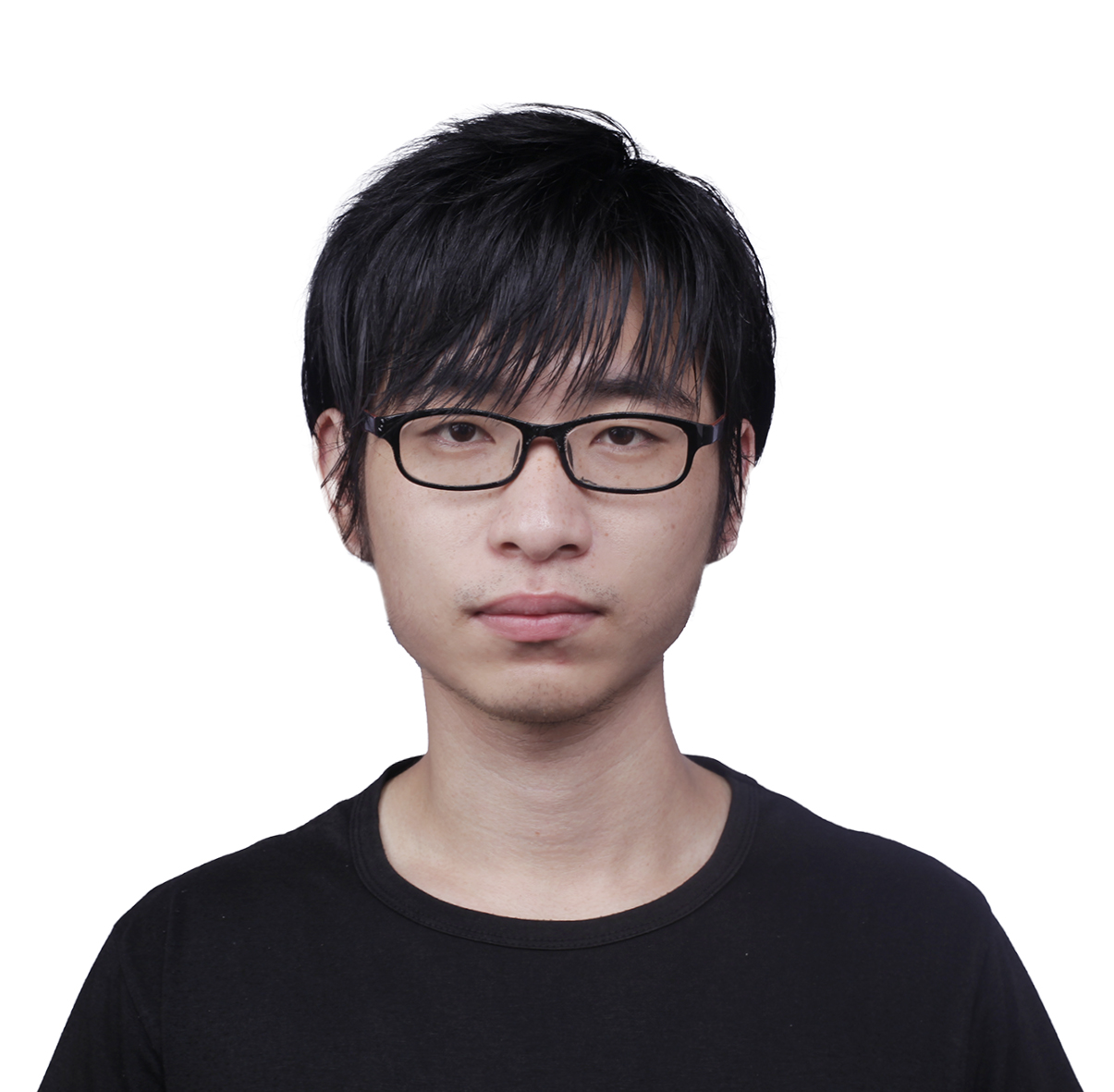}}]{Guo-Hua Wang}
	received  his  BS  degree  in the School of Management and Engineering from Nanjing University. He is currently a Ph.D. student in the Department of Computer Science and Technology in Nanjing University, China. His research interests are computer vision and machine learning.
\end{IEEEbiography}

\begin{IEEEbiography}[{\includegraphics[width=1in,height=1.25in,clip,keepaspectratio]{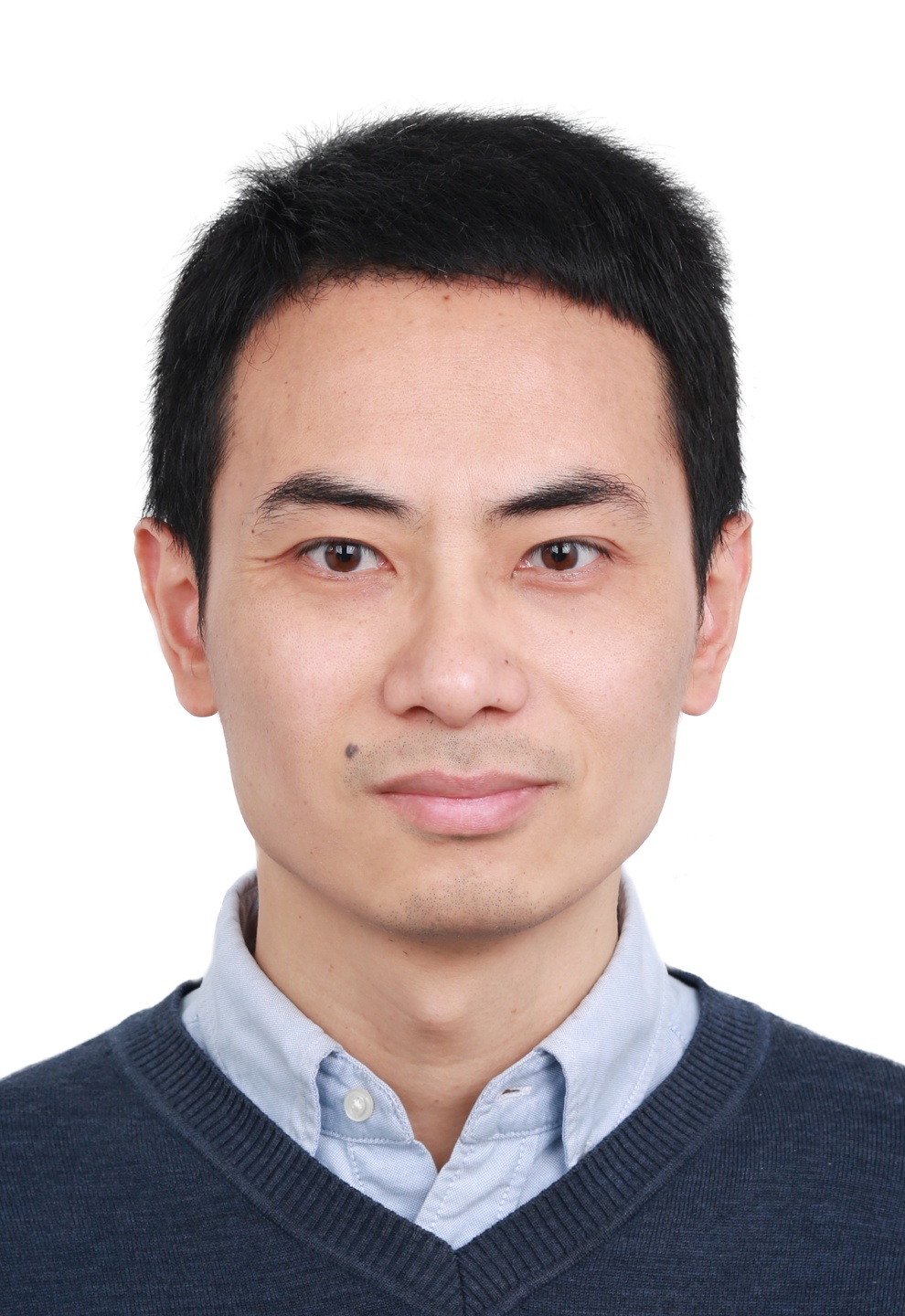}}]{Jianxin Wu}
	(Member, IEEE) received the BS and MS degrees from Nanjing University, and the PhD degree from the Georgia Institute of Technology, all in computer science. He is currently a professor with the Department of Computer Science and Technology and School of Artificial Intelligence, Nanjing University, China, and associated with the State Key Laboratory for Novel Software Technology, China. His research interests include computer vision and machine learning. He was the (senior) area chair of CVPR, ICCV, ECCV, AAAI, and IJCAI and an associate editor for the IEEE Transactions on Pattern Analysis and Machine Intelligence.
\end{IEEEbiography}

\clearpage
\newpage

\appendices
\section{Proof of Claim~\ref{claim:upper-bound}}
\label{app:proof:claim1}
We omit the parameters $\theta$ in $q_\theta(y|x)$ and $q_\theta(y|f,x)$ for simplicity.
\begin{align}
&\phantom{==}  KL\left[p(y, f|x)||q(y, f|x)\right]\\
&= \iint p(y,f|x)\ln\frac{p(y,f|x)}{q(y, f|x)}dydf\\
&= \iint p(f|y,x)p(y|x)\ln\frac{p(f|y, x)p(y|x)}{q(f|y, x)q(y|x)}dydf\\
&= \int p(y|x) \int p(f|y,x)\left[\ln\frac{p(y|x)}{q(y|x)} + \ln\frac{p(f|y, x)}{q(f|y, x)}\right]dydf\\
&= \int p(y|x) \int p(f|y,x)\ln\frac{p(y|x)}{q(y|x)}dydf \notag \\
&\qquad +\int p(y|x) \int p(f|y,x)\ln\frac{p(f|y, x)}{q(f|y, x)}dydf\\
&= \int p(y|x) \ln\frac{p(y|x)}{q(y|x)}dy \notag \\
&\qquad +\int p(y|x) KL\left[p(f|y,x)||q(f|y, x)\right]dy\\
&= KL\left[p(y|x)||q(y|x)\right] \notag \\
&\qquad + \int p(y|x) KL\left[p(f|y,x)||q(f|y, x)\right]dy\\
&\ge KL\left[p(y|x)||q(y|x)\right]\,.
\end{align}

\section{Proof of Claim~\ref{claim:loss-decouple}}
\label{app:proof:claim2}
We omit the parameters $\theta$ in $q_\theta(y|x)$ and $q_\theta(y|f,x)$ for simplicity.
\begin{align}
&\phantom{==} KL\left[p(y, f|x)||q(y, f|x)\right]\\
&= \iint p(y,f|x)\ln\frac{p(y,f|x)}{q(y, f|x)}dydf \\
&= \iint p(y|f,x)p(f|x)\ln\frac{p(y|f, x)p(f|x)}{q(y, f|x)}dydf\\
&= \int p(f|x) \int p(y|f,x)\ln\frac{p(y|f, x)p(f|x)}{q(y, f|x)}dydf\\
&= \mathbb{E}_{f\sim p(f|x)} \int p(y|f,x)\ln\frac{p(y|f, x)p(f|x)}{q(y, f|x)}dy\\
&= \mathbb{E}_{f\sim p(f|x)} \left[\int p(y|f,x)\ln p(f|x)dy \right. \notag \\
&\qquad + \left.\int p(y|f,x)\ln\frac{p(y|f, x)}{q(y, f|x)}dy\right]\\
&= \mathbb{E}_{f\sim p(f|x)} \left[\ln p(f|x) + \int p(y|f,x)\ln\frac{p(y|f, x)}{q(y|f,x)q(f|x)}dy\right]\\
&= \mathbb{E}_{f\sim p(f|x)} \left[\ln p(f|x) - \int p(y|f,x)\ln q(f|x)dy \right.\notag \\
&\qquad + \left.\int p(y|f,x)\ln\frac{p(y|f, x)}{q(y|f,x)}dy\right]\\
%&= \underset{f\sim p(f|x)}{\mathbb{E}} \left\{\ln p(f|x) - \ln q(f|x) + KL\left[p(y|f,x) || q(y|f,x)\right]\right\}\\
&=\underset{f\sim p(f|x)}{\mathbb{E}} \left\{- \ln q(f|x) + KL\left[p(y|f,x) || q(y|f,x)\right]\right\}+C\,.
\end{align}

For feature mimicking, we let $q_\theta(f|x)$ be a Gaussian whose distribution parameters are computed from $x$ with a deep neural network. So we have
\begin{equation}
   \label{eq:FM-loss}
   q_\theta(f|x) = \frac{1}{\sqrt{2\pi} \sigma_\theta(x)}\exp\left(-\frac{1}{2}\left(\frac{f-\mu_\theta(x)}{\sigma_\theta(x)}\right)^2\right)\,.
\end{equation}
Hence, the feature mimicking loss becomes
\begin{align}
&\phantom{=\;\;} \mathbb{E}_{f\sim p(f|x)} [- \ln q_\theta(f|x)] \\
&= \mathbb{E}_{f\sim p(f|x)} \left[-\ln \frac{1}{\sqrt{2\pi} \sigma_\theta(x)}\exp\left(-\frac{1}{2}\left(\frac{f-\mu_\theta(x)}{\sigma_\theta(x)}\right)^2\right)\right]\\
&= \mathbb{E}_{f\sim p(f|x)} \left[\frac{1}{2}\left(\frac{f-\mu_\theta(x)}{\sigma_\theta(x)}\right)^2 -\ln \frac{1}{\sqrt{2\pi} \sigma_\theta(x)}\right]\,.
\end{align}
If we assume the variance is a constant, and let $\sigma(x)=\frac{1}{\sqrt{2\beta}}$, then we have
\begin{align}
&\phantom{=\;\;} \mathbb{E}_{f\sim p(f|x)} [- \ln q_\theta(f|x)] \\
&= \mathbb{E}_{f\sim p(f|x)} \left[\beta\left(f-\mu_\theta(x)\right)^2 -\ln \frac{\sqrt{2\beta}}{\sqrt{2\pi}}\right]\\
&= \mathbb{E}_{f\sim p(f|x)} \left[\beta\left(f-\mu_\theta(x)\right)^2 \right]+C \,.
\end{align}

\section{Proof of Claim~\ref{claim:variance}}
\label{app:proof:claim3}

Given $n$ training samples $x_1,x_2,\cdots,x_n$ and the loss function 
\begin{equation}
  \mathcal{L} = -\ln q_\theta (f|x)\,,
\end{equation}
we can obtain local minima solutions by finetuning. By Conjecture~\ref{conj:convexity_of_finetune}, the loss landscape is convex w.r.t. parameters, and the local minima will not escape from the current optimization basin. Hence, the solution can be derived by 
\begin{equation}
  \argmax_{\theta} \sum_{i=1}^{n}\ln q_{\theta}(f_i|x_i)\,.
\end{equation}
Note that it is equivalent to a Maximum Likelihood Estimate (MLE). The log-likelihood for each sample is 
\begin{equation}
   l_i(\theta) = \ln q(f_i|x_i)\,,
\end{equation}
and the total log-likelihood is 
\begin{equation}
   l(\theta) = \sum_{i=1}^{n}l_i(\theta)\,.
\end{equation}
For MLE, the condition for achieving optimal solutions $\hat{\theta}$ is $\frac{\partial l(\theta)}{\partial \theta}|_{\theta=\hat{\theta}}=0$. With the Taylor expansion on $\tilde{\theta}$ and omitting high order terms,
\begin{equation}
   0= \frac{\partial l(\theta)}{\partial \theta}|_{\theta=\hat{\theta}}\approx \frac{\partial l(\tilde{\theta})}{\partial \tilde{\theta}}+\frac{\partial^2 l(\tilde{\theta})}{\partial \tilde{\theta}^2}(\hat{\theta}-\tilde{\theta})\,.
\end{equation}
Hence, we have 
\begin{equation}
  \label{eq:solutions}
   \hat{\theta}\doteq \tilde{\theta} - \frac{\frac{\partial l(\tilde{\theta})}{\partial \tilde{\theta}}}{\frac{\partial^2 l(\tilde{\theta})}{\partial \tilde{\theta}^2}}\,.
\end{equation}
Note that this equation holds approximately under our Conjecture~\ref{conj:convexity_of_finetune}. For simplicity, we denote $\tilde{\theta}$ as $\theta$ in the rest of this proof. Next, we will analyze the variance of $\hat{\theta}$. For each sample, 
\begin{equation}
   \frac{\partial l_i(\theta)}{\partial \theta}=\frac{\dot{q}_{\theta}(f_i|x_i)}{q_{\theta}(f_i|x_i)}\,,
\end{equation}
where $\dot{q}_{\theta}(f_i|x_i)$ denotes $\frac{\partial q_{\theta}(f_i|x_i)}{\partial \theta}$. 
\begin{align}
\mathbb{E}_{f\sim q(f|x_i)}\left[\frac{\partial l_i(\theta)}{\partial \theta}\right]
&=\int \frac{\dot{q}_{\theta}(f|x_i)}{q_{\theta}(f|x_i)}q_{\theta}(f|x_i)df\\
&=\int \dot{q}_{\theta}(f|x_i)df \\
&=\frac{\partial}{\partial \theta}\int q_{\theta}(f|x_i)df \\
&=\frac{\partial 1}{\partial \theta}=0\,.
\end{align}
\begin{equation}
   Var_{f\sim q(f|x_i)}\left[\frac{\partial l_i(\theta)}{\partial \theta}\right]=\int \dot{l}_i(\theta)^2q_{\theta}(f|x_i)df\,.
\end{equation}
The Fisher information is $\mathcal{I}_i(\theta)=\int \dot{l}_i(\theta)^2q_{\theta}(f|x_i)df$, and 
\begin{equation}
   \mathcal{I}(\theta)=\mathbb{E}_{x\sim p(x)}\int \dot{l}_i(\theta)^2q_{\theta}(f|x)df\,.
\end{equation}
Note that $\frac{\dot{l}(\theta)}{n}=\frac{1}{n}\sum_{i=1}^{n}\dot{l}_i(\theta)$, and with the central limit theorem (CLT), we have
\begin{equation}
   \label{eq:l1}
   \frac{\dot{l}(\theta)}{n}\sim \mathcal{N}(0, \frac{\mathcal{I}(\theta)}{n})\,.
\end{equation}

For $\frac{\partial^2 l_i(\theta)}{\partial \theta^2}$, we have
\begin{equation}
   \frac{\partial^2 l_i(\theta)}{\partial \theta^2}=\frac{\ddot{q}_\theta(f_i|x_i)}{q_{\theta}(f_i|x_i)}-\left(\frac{\dot{q}_{\theta}(f_i|x_i)}{q_{\theta}(f_i|x_i)}\right)^2\,.
\end{equation}
Then,
\begin{align}
&\phantom{=\;\;}\mathbb{E}_{f\sim q(f|x_i)}\left[\frac{\partial^2 l_i(\theta)}{\partial \theta^2}\right] \\
&=\mathbb{E}_{f\sim q(f|x_i)}\left[\frac{\ddot{q}_\theta(f|x_i)}{q_{\theta}(f|x_i)}\right]-\mathbb{E}_{f\sim q(f|x_i)}\left(\frac{\dot{q}_{\theta}(f|x_i)}{q_{\theta}(f|x_i)}\right)^2 \\
&=-\mathcal{I}_i(\theta)\,.
\end{align}
With the law of large numbers, we have
\begin{equation}
   \label{eq:l2}
   \frac{\ddot{l}(\theta)}{n}=\frac{1}{n}\sum_{i=1}^{n}\ddot{l}_i(\theta)\overset{P}{\rightarrow} -\mathcal{I}(\theta)\,.
\end{equation}
With Eqs.~\ref{eq:l1},~\ref{eq:l2} and~\ref{eq:solutions}, we have 
\begin{equation}
   \hat{\theta}\sim \mathcal{N}(\tilde{\theta}, \frac{1}{n\mathcal{I}(\tilde{\theta})})\,.
\end{equation}

\section{Proof of Claim~\ref{claim:FM:variance}}
\label{app:proof:claim4}

To finetune with the loss in Eq.~\ref{eq:MSE_loss}, we have assumed $q(f|x)$ is a Gaussian and $\sigma(x)=\frac{1}{\sqrt{2\beta}}$, hence
\begin{equation}
   q(f|x) = \sqrt{\frac{\beta}{\pi}}\exp\left(-\beta\left(f-\mu_\theta(x)\right)^2\right)\,.
\end{equation}
We assume $\mu(x)=w\cdot x + b$, where $w$ and $b$ are parameters $\theta$. First, we consider $w$:
\begin{equation}
   \dot{q}_w(f|x) = \sqrt{\frac{\beta}{\pi}}\exp\left(-\beta\left(f-\mu(x)\right)^2\right)2\beta\left(f-\mu(x)\right)x\,.
\end{equation}
And,
\begin{align}
   \dot{l}_i(w)&=\frac{\dot{q}_w(f_i|x_i)}{q_w(f_i|x_i)} \\
     &=2\beta\left(f_i-\mu(x_i)\right)x_i\\
	 &=2\beta x_i(f_i-w\cdot x_i-b)\,,\\
   \ddot{l}_i(w)&=-2\beta x_i^2\,.
\end{align}
\begin{equation}
   \mathcal{I}(w)=-\mathbb{E}_{x}\ddot{l}_i(w)=2\beta \mathbb{E}[x^2]\,.
\end{equation}
By Claim~\ref{claim:variance}, we achieve
\begin{equation}
   \hat{w}\sim \mathcal{N}(\tilde{w}, \frac{1}{2n\beta \mathbb{E}[x^2]})\,.
\end{equation}

Then, for $b$, we also have
\begin{equation}
  \dot{q}_b(f|x) = \sqrt{\frac{\beta}{\pi}}\exp\left(-\beta\left(f-\mu(x)\right)^2\right)2\beta\left(f-\mu(x)\right)\,.
\end{equation}
\begin{equation}
  \dot{l}_i(b)=\frac{\dot{q}_b(f_i|x_i)}{q_b(f_i|x_i)}=2\beta\left(f_i-\mu(x_i)\right)=2\beta(f_i-w\cdot x_i-b)\,.
\end{equation}
\begin{equation}
  \ddot{l}_i(b)=-2\beta\,.
\end{equation}
\begin{equation}
  \mathcal{I}(b)=-\mathbb{E}_{x}\ddot{l}_i(b)=2\beta\,.
\end{equation}
By Claim~\ref{claim:variance}, we achieve
\begin{equation}
  \hat{b}\sim \mathcal{N}(\tilde{b}, \frac{1}{2n\beta})\,.
\end{equation}

\section{Proof of Claim~\ref{claim:class:variance}}
\label{app:proof:claim5}

The classification loss in Eq.~\ref{eq:vanilla} can be written into 
\begin{equation}
  \mathbb{E}_{x\sim p(x)} KL\left[p(y|x) || q(y|x)\right] \,.
\end{equation}
With the one-hot coding $\vy$, this loss is equal to 
\begin{align}
   \label{eq:class}
KL\left[p(y|x) || q(y|x)\right]
&=\sum_{j=1}^{C}p(y_j|x)\ln\frac{p(y_j|x)}{q(y_j|x)} \\
&=-\ln q(y_c|x)\,,
\end{align}
where $c$ denotes the ground-truth class. This loss follows the canonical form in Claim~\ref{claim:variance}. In a deep neural network, $x$ is first transformed into the feature $f$, then $q(y_c|f)$ is calculated as
\begin{equation}
   q(y_c|f)=\sigma(\vw\cdot f + \vb)_c=\frac{\exp(w_c\cdot f+b_c)}{\sum_{j=1}^{C}\exp(w_j\cdot f + b_j)}\,.
\end{equation}
If the $i$-th sample belongs to class $c$, the log-likelihood is 
\begin{align}
   l_i(\vw, \vb) &= \ln\sigma(\vw\cdot f_i + \vb)_c  \\
     &=w_c\cdot f_i+b_c-\ln\sum_{j=1}^{C}\exp(w_j\cdot f_i + b_j)\,.
\end{align}
First, we consider $\vw$. We have
\begin{equation}
   \dot{l}_i(w_c) = f_i- \frac{\exp(w_c\cdot f_i+b_c)f_i}{\sum_{j=1}^{C}\exp(w_j\cdot f_i + b_j)}=f_i-\sigma(\vw\cdot f_i + \vb)_cf_i\,,
\end{equation}
\begin{equation}
   \dot{l}_i(w_{c'}) = -\frac{\exp(w_{c'}\cdot f_i+b_{c'})f_i}{\sum_{j=1}^{C}\exp(w_j\cdot f_i + b_j)}=-\sigma(\vw\cdot f_i + \vb)_{c'}f_i\,,
\end{equation}
where $c'\neq c$. Then,
\begin{align}
   \ddot{l}_i(w_{j}) &= \frac{\partial \sigma(\vw f_i + \vb)_{j}f_i}{\partial w_j} \\
   &= -\sigma(\vw f_i + \vb)_j[1-\sigma(\vw f_i + \vb)_j]f_i^2\,.
\end{align}
If we denote $\epsilon_j=\sigma(\vw\cdot f_i + \vb)_j[1-\sigma(\vw\cdot f_i + \vb)_j]$, then $\epsilon_j$ always approach zero in practice. We have $\ddot{l}_i(w_{j})=-\epsilon_j f_i^2$, and 
\begin{equation}
   \mathcal{I}(w_j)=-\mathbb{E}_f\ddot{l}_i(w_j)=\epsilon_j\mathbb{E}[f^2]\,.
\end{equation}

Finally, we achieve
\begin{equation}
   \hat{w}_j\sim \mathcal{N}(\tilde{w}_j, \frac{1}{n\epsilon_j\mathbb{E}[f^2]})\,.
\end{equation}

Now, we consider $\vb$.
\begin{equation}
  \dot{l}_i(b_c) = 1- \frac{\exp(w_c\cdot f_i+b_c)}{\sum_{j=1}^{C}\exp(w_j\cdot f_i + b_j)}=1-\sigma(\vw\cdot f_i + \vb)_c\,,
\end{equation}
\begin{equation}
  \dot{l}_i(b_{c'}) = -\frac{\exp(w_{c'}\cdot f_i+b_{c'})}{\sum_{j=1}^{C}\exp(w_j\cdot f_i + b_j)}=-\sigma(\vw\cdot f_i + \vb)_{c'}\,,
\end{equation}
where $c'\neq c$. And,
\begin{equation}
  \ddot{l}_i(b_{j}) = \frac{\partial \sigma(\vw f_i + \vb)_{j}}{\partial b_j}=-\sigma(\vw f_i + \vb)_j[1-\sigma(\vw f_i + \vb)_j]\,.
\end{equation}
\begin{equation}
  \mathcal{I}(b_j)=-\mathbb{E}_f\ddot{l}_i(b_j)=\epsilon_j\,.
\end{equation}

Finally, we achieve
\begin{equation}
  \hat{b}_j\sim \mathcal{N}(\tilde{b}_j, \frac{1}{n\epsilon_j})\,.
\end{equation}

\end{document}